\pdfoutput=1

\documentclass[11pt]{article}

\usepackage[final]{acl}

\usepackage{times}
\usepackage{latexsym}

\usepackage[T1]{fontenc}

\usepackage[utf8]{inputenc}

\usepackage{microtype}

\usepackage{inconsolata}

\usepackage{graphicx}
\usepackage{hyperref}       
\usepackage{url}            
\usepackage{booktabs}       
\usepackage{arydshln}
\usepackage{amsfonts}       
\usepackage{nicefrac}       
\usepackage{microtype}      

\usepackage{xcolor}         
\usepackage{marvosym}
\usepackage{graphicx}
\usepackage{colortbl}
\usepackage{multirow}
\usepackage{array}
\usepackage{tcolorbox}
\usepackage{subcaption}
\usepackage{algorithm}
\usepackage{algpseudocode}
\usepackage{amsmath}

\title{CRVQ: Channel-Relaxed Vector Quantization\\
for Extreme Compression of LLMs}

\author{First Author \\
  Affiliation / Address line 1 \\
  Affiliation / Address line 2 \\
  Affiliation / Address line 3 \\
  \texttt{email@domain} \\\And
  Second Author \\
  Affiliation / Address line 1 \\
  Affiliation / Address line 2 \\
  Affiliation / Address line 3 \\
  \texttt{email@domain} \\}

\author{
Yuzhuang Xu \quad Shiyu Ji \quad Qingfu Zhu \quad Wanxiang Che\textsuperscript{\Letter} \\
  Research Center for Social Computing and Interactive Robotics (SCIR)\\
Harbin Institute of Technology, Harbin, China\\
  \texttt{\{xyz, car\}@ir.hit.edu.cn} \\
}

\begin{document}
\maketitle
\begin{abstract}
Powerful large language models (LLMs) are increasingly expected to be deployed with lower computational costs, enabling their capabilities on resource-constrained devices. Post-training quantization (PTQ) has emerged as a star approach to achieve this ambition, with best methods compressing weights to less than 2 bit on average. In this paper, we propose \textbf{C}hannel-\textbf{R}elaxed \textbf{V}ector \textbf{Q}uantization (CRVQ), a novel technique that significantly improves the performance of PTQ baselines at the cost of only minimal additional bits. This state-of-the-art extreme compression method achieves its results through two key innovations: (1) carefully selecting and reordering a very small subset of critical weight channels, and (2) leveraging extended codebooks to relax the constraint of critical channels. With our method, we demonstrate a 38.9\% improvement over the current strongest sub-2-bit PTQ baseline, enabling nearer lossless 1-bit compression. Furthermore, our approach offers flexible customization of quantization bit-width and performance, providing a wider range of deployment options for diverse hardware platforms.
\end{abstract}

\section{Introduction}
\label{sec:intro}

Large language models (LLMs) have made remarkable strides, with open-source models like LLaMA~\citep{llama2023, llama2023se}, Phi~\citep{phi2024} and Qwen~\citep{qwen2023, qwen2024se} achieving exceptional performance. 
However, their massive sizes present challenges for deployment and usability. For instance, the widely adopted and powerful LLaMA2-70B model requires over 140GB when loaded in FP16 format, which makes application on most single-GPUs infeasible. This computational and memory overhead becomes even more prohibitive for resource-constrained devices. As the demand for edge-device LLM inference grows, especially on mobile platforms~\citep{minicpm2024, phi2024}, researchers have turned to model compression techniques~\citep{gptq2022, omniquant2023, onebit2024} to enable efficient deployment without compromising performance noticeably.

\begin{figure}[t]
  \centering
  \begin{subfigure}[b]{\columnwidth}
    \centering
        \includegraphics[width=1.0\textwidth]{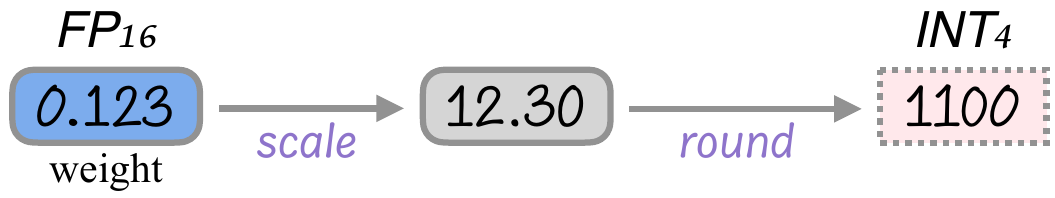}
        \vspace{-0.7cm}
        \caption{uniform quantization}
        \label{fig:intro1}
        \vspace{0.2cm}
    \end{subfigure}
  \begin{subfigure}[b]{\columnwidth}
    \centering
        \includegraphics[width=1.0\textwidth]{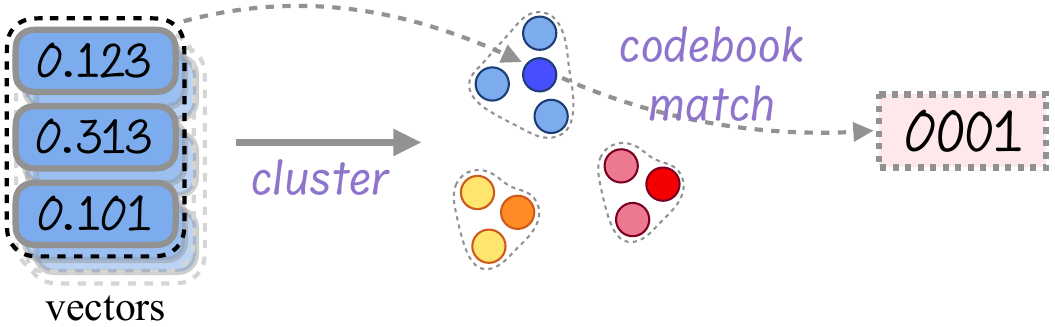}
        \vspace{-0.7cm}
        \caption{vector quantization}
        \label{fig:intro2}
        \vspace{0.1cm}
    \end{subfigure}
  \begin{subfigure}[b]{\columnwidth}
    \centering
        \includegraphics[width=1.0\textwidth]{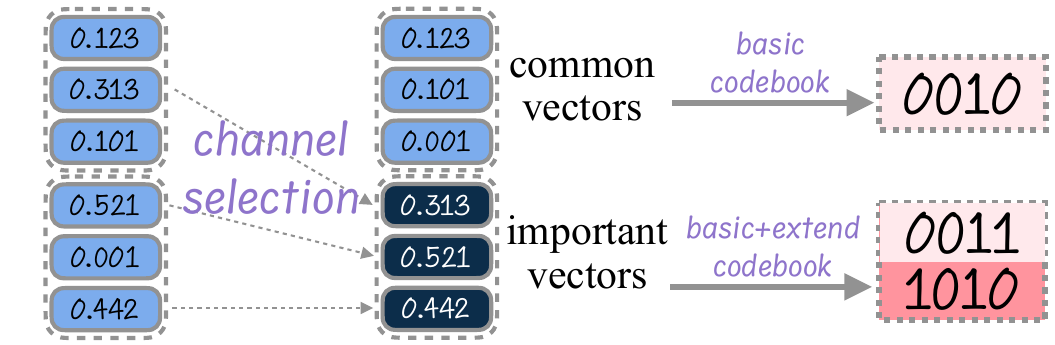}
        \vspace{-0.5cm}
        \caption{CRVQ}
        \label{fig:intro3}
        \vspace{-0.2cm}
    \end{subfigure}
  \caption{Comparison of different quantization methods. Uniform quantization treats each weight to be quantized as a scalar, whereas vector quantization considers weight segments as vectors. Our proposed CRVQ introduces distinct importance across weight channels, with the more critical channels selected and highlighted in dark colors. We use additional extended codebooks to quantize the vectors formed by these critical channels.}
\label{fig:intro}
\vspace{-1em}
\end{figure}


Post-Training Quantization (PTQ) efficiently compresses LLMs by converting pre-trained models to lower-bit formats with relatively few computations~\citep{gptq2022, quip2024, billm2024}. Its effectiveness has driven broad research and practical adoption. Notably, recent advances achieve 3-bit compression while astonishingly maintaining lossless performance~\citep{quip2024}. 
Unfortunately, the performance of powerful PTQ methods rapidly deteriorates in extremely low bit-widths scenarios. For instance, GPTQ~\citep{gptq2022}, the most popular method for INT4 quantization, completely fails when weights are compressed to 2-bit. 
A possible reason is that these methods treat each weight as an isolated and uniformly distributed element, independently quantizing them while ignoring the inherent patterns among weights~\citep{aqlm2024, quip2024}. This oversight results in suboptimal bit utilization, causing these methods to quickly reach their performance limits. Recently, researchers turn to vector quantization (VQ) to address this challenges in PTQ~\citep{quip2024fe, quip2024, aqlm2024}. By treating weights as vectors rather than isolated scalars (see Figure~\ref{fig:intro1} and~\ref{fig:intro2}), VQ-based approaches appear to show promising results. For example, 
AQLM~\citep{aqlm2024} enables seamless adoption of 1-bit settings while surpassing previous strong baselines~\citep{pbllm2024, billm2024}. 
However, noticeable performance degradation of VQ-based methods remains unsatisfactory at 1-bit level.

Inspired by LLM.int8()~\citep{llmint2022} and SpQR~\citep{spqr2023}, we propose \textbf{C}hannel-\textbf{R}elaxed \textbf{V}ector \textbf{Q}uantization (CRVQ) in this paper, aiming to treat different weight channels with varying importance in VQ for better quantization. Traditional VQ generally assume equal importance across all weight channels~\citep{quip2024, aqlm2024}. In contrast, we argue that even in VQ, a small subset of critical channels can play a pivotal role in maintaining model performance. CRVQ enables these critical channels to break the limits of the current quantization level, see the comparison in Figure~\ref{fig:intro}. This involves two simple yet effective steps: First, we explore useful criteria for evaluating the importance of channels and reorder them, grouping channels of similar importance closer. Then, a few critical channels are selected and quantized with additional codebooks to enhance their representation. 
CRVQ significantly improves 1-bit PTQ, reducing perplexity by 38.9\% and boosting zero-shot accuracy by 12.9\% on LLaMA2-7B, with only a negligible bit-width cost. 
In summary, our contributions are 3-fold:

\begin{itemize}
    \item We propose CRVQ, a novel and effective 1-bit PTQ method that selects critical weight channels and applies extended codebook fitting. Punching above its weight, it excels in extreme compression.
    \item Based on an in-depth analysis of the factors influencing performance, we highlight that CRVQ is a more efficient and flexible solution for achieving superior quantization performance on resource-constrained devices compared to traditional VQ.
    \item Extensive experiments demonstrate that our approach performs well in different models of varying sizes, ranging from 1.3B to 13B. On LLaMA2-7B, it reduces perplexity by 39\% over AQLM with only a 0.06-bit overhead, showcasing its effectiveness.
\end{itemize}

\section{Related Work}
\label{sec:work}


Similar to model pruning~\citep{sparsegpt2023, wanda2023} and knowledge distillation~\citep{gkd2023, distilling2023}, model quantization is a key technique for compressing LLMs and reducing computational overhead. 
It can be classified into quantization-aware training (QAT) and post-training quantization (PTQ) based on its integrating stage.

QAT integrates quantization into the intermediate computations of model training, reducing performance degradation caused by low precision and leading superior results~\citep{llmqat2023,qlora2023,efqat2024,onebit2024,bmos2024}. While effective for extreme quantization, it still suffers from certain performance limitations and high computational costs. As this paper primarily focuses on PTQ, the remainder of this section will concentrate on the features and representative methods of PTQ.


PTQ transforms LLMs into their low-bit counterparts using solvers and limited calibration data, without requiring extensive training. On the one hand, PTQ has made remarkable progress in extreme compression scenarios. GPTQ~\citep{gptq2022} iteratively quantizes columns while adjusting other weights to preserve accuracy, achieving 4-bit quantization. QuIP\#~\citep{quip2024} employs compact E8P codebooks and incoherence processing to enable high-precision 2-bit VQ. Similarly, AQLM~\citep{aqlm2024} 
leverages learnable codebooks that can effectively adapt to the 1-bit setting. 
Additionally, BiLLM~\citep{billm2024} achieves binarization by combining weight selection and residual approximation. 
On the other hand, the idea of mixed precision has been widely applied in quantization research. LLM.int8()~\citep{llmint2022} is an early approach that proposed mixed-precision decomposition to preserve the functionality of critical channels. Other notable works employing special handling of outlier channels include SmoothQuant~\citep{smooth2023}, AWQ~\citep{awq2023} and SpQR~\citep{spqr2023}, etc. Our work pushes the boundaries of 1-bit PTQ by enhancing VQ. Unlike existing methods, we are the first to explore the varying importance of channels within VQ, even though these channels are not adjacent. This novel perspective sets our approach apart from prior work.

\section{Background}
\label{sec:back}

\subsection{Vector Quantization}


VQ partitions the weight matrix into vectors and uses the vector as the basic quantization unit. 
Let $\mathcal{S}=\{\mathbf{w}_i \in \mathbb{R}^d\}$ represent the set of \textit{d}-dimensional vectors obtained by partitioning $\mathbf{W}^{M\times N}$. 
\begin{align}
    \mathcal{S} = \left\{ \mathbf{W}_{i, j:j+d} \mid i \in [\ 0,M\ ), j \in [\ 0,N\ ) \right. ,\notag \\
    \left. j \bmod d = 0 \right\}.
\end{align}
VQ applies the \textit{K}-means algorithm to return the cluster centers $\mathcal{C} = \{\mathbf{c}_i \in \mathbb{R}^d\}$, which are subsequently used as the codebook:
\begin{equation}
\label{eq:kmeans}
    \mathcal{C}=K\textrm{-means}\left(\mathcal{S}\right).
\end{equation}
If each vector in $\mathcal{C}$ is encoded using $e$-bit binary number, the codebook $\mathcal{C}$ contains $2^e$ vectors, i.e. $|\mathcal{C}|=2^e$. After the codebook is built, each vector in $\mathcal{S}$ is replaced with its closest vector from $\mathcal{C}$. 
\begin{equation}
\mathcal{S}\approx\left\{\hat{\mathbf{w}}_i = \arg\min_{\mathbf{c}_j \in \mathcal{C}} \|\mathbf{w}_i - \mathbf{c}_j\|_2\right\}.
\end{equation}
Now, the original $d$-dimensional vector $\mathbf{w}$ is encoded into an $e$-bit binary code $q$ as follows: 
\begin{equation}
q_i = \arg\min_{j} \|\mathbf{w}_i - \mathbf{c}_j\|_2, \quad q_i \in {0, 1, \dots, 2^e - 1}.
\end{equation}
Naturally, the capacity of the codebook increases with $e$, leading to better approximation of $\mathbf{W}$. 
To reduce approximation error, we can iteratively refine the process above by quantizing the residual $\Delta\mathbf{W}=\mathbf{W}-\hat{\mathbf{W}}$, introducing more codebooks and codes. Beam search is typically used for the optimal multi-step encoding~\citep{aqlm2024}.

To further mitigate the model sensitivity to extreme compression, block or end-to-end fine-tuning a few parameters is considered highly beneficial~\citep{quip2024, aqlm2024}.
\begin{equation}
\mathcal{L}=\|\mathrm{block}(\mathbf{X})-\mathrm{quantized\_block}(\mathbf{X})\|_{2}.
\end{equation}
Let $\mathcal{L}$ be the loss and $\mathbf{X}$ the block input here. While end-to-end fine-tuning slows PTQ, it remains more efficient than most QAT methods.


\subsection{From AQLM to 1-bit Quantization}
\label{subsec:aqlm}

\begin{figure}[h]
    \centering
    \begin{subfigure}[b]{\columnwidth}
    \centering
        \includegraphics[width=0.9\textwidth]{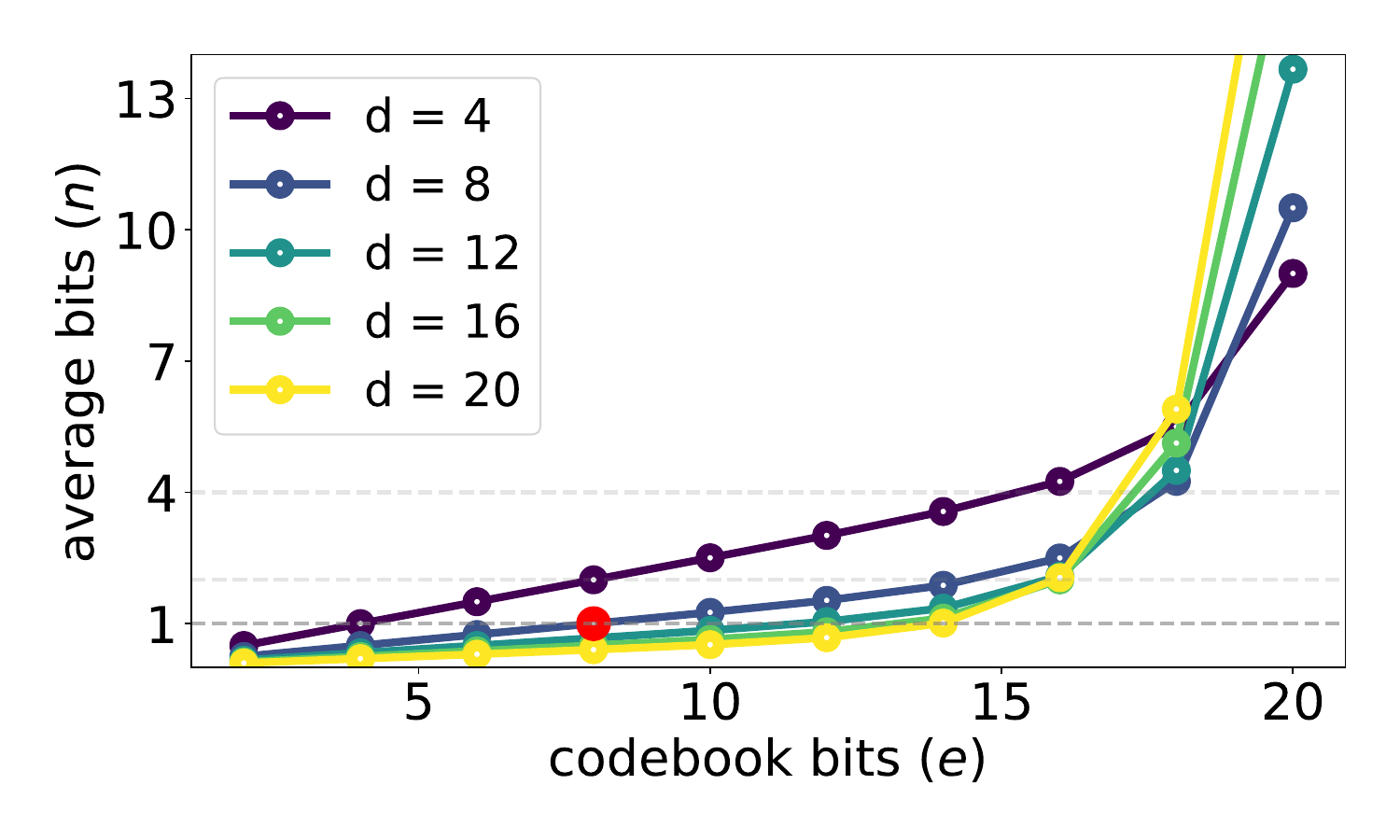}
        \caption{$m=1$}
        \label{fig:m1}
        \vspace{0.2cm}
    \end{subfigure}
    \begin{subfigure}[b]{\columnwidth}
    \centering
        \includegraphics[width=0.9\textwidth]{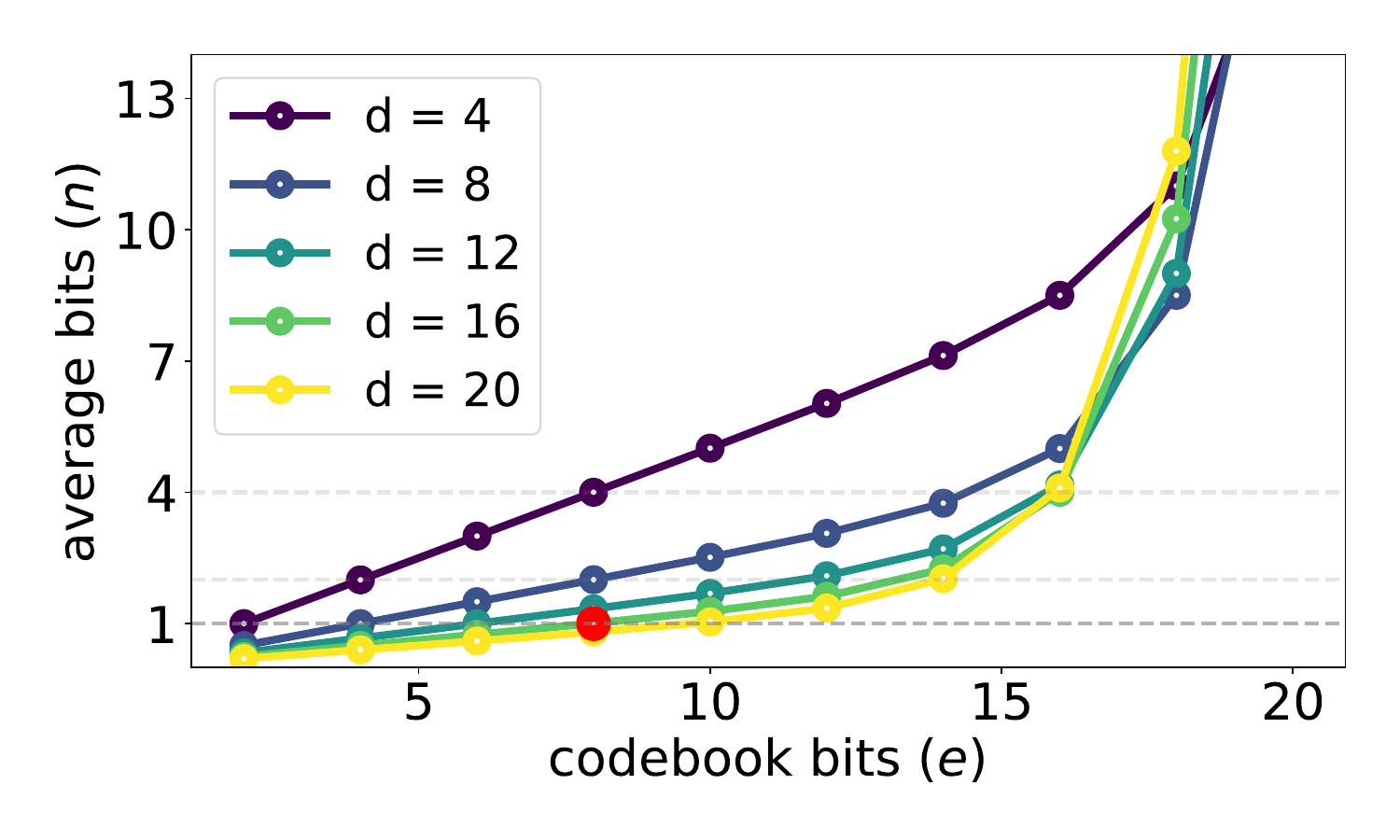}
        \caption{$m=2$}
        \label{fig:m2}
    \end{subfigure}
    \vspace{-0.6cm}
    \caption{Visualization of Equation~\ref{eq:bitwidth}. The curves represent bit-width based on the vector dimension $d$ and the codebook bit-width $e$. The top (a) and bottom (b) plots correspond to $m=1$ and $m=2$, respectively.}
    \label{fig:bits}
\end{figure}

Compared to QuIP\#, AQLM are more adaptable for the 1-bit quantization level, as it avoids the challenge of relying on predefined dense codebooks. Let the number of codebooks is $m$, the average bit-width $n$ of a quantized $4096 \times 4096$ matrix can be calculated as follows (see Section~\ref{app:bits} for details):
\begin{equation}
\label{eq:bitwidth}
    n = \frac{me}{d} + 2^{e-20}md.
\end{equation}
1-bit quantized models can be built by selecting different combinations of hyper parameters. Figure~\ref{fig:bits} illustrates the bit-width curve for various $e,d$ corresponding to two different codebook number $m$, respectively. It can be observed that when the codebook bit-width $e$ exceeds 14, the average bit-width rises sharply. It is caused by the excessive codebook storage overhead, reducing the proportion of bit-width allocated to weights. Consequently, larger values of $e$ are avoided.

We adopt a commonly used value of $e = 8$, which limits the feasible settings for 1-bit quantization to $m = 1, d = 8$ or $m = 2, d = 16$. In this work, we choose the configuration with a smaller bit-width, $m = 1, d = 8, e = 8$. As we will demonstrate in Section~\ref{subsec:combine}, CRVQ also performs as expected under the $m = 2, d = 16$ setting.

\begin{figure*}[htbp]
  \centering
  \includegraphics[width=0.99\textwidth]{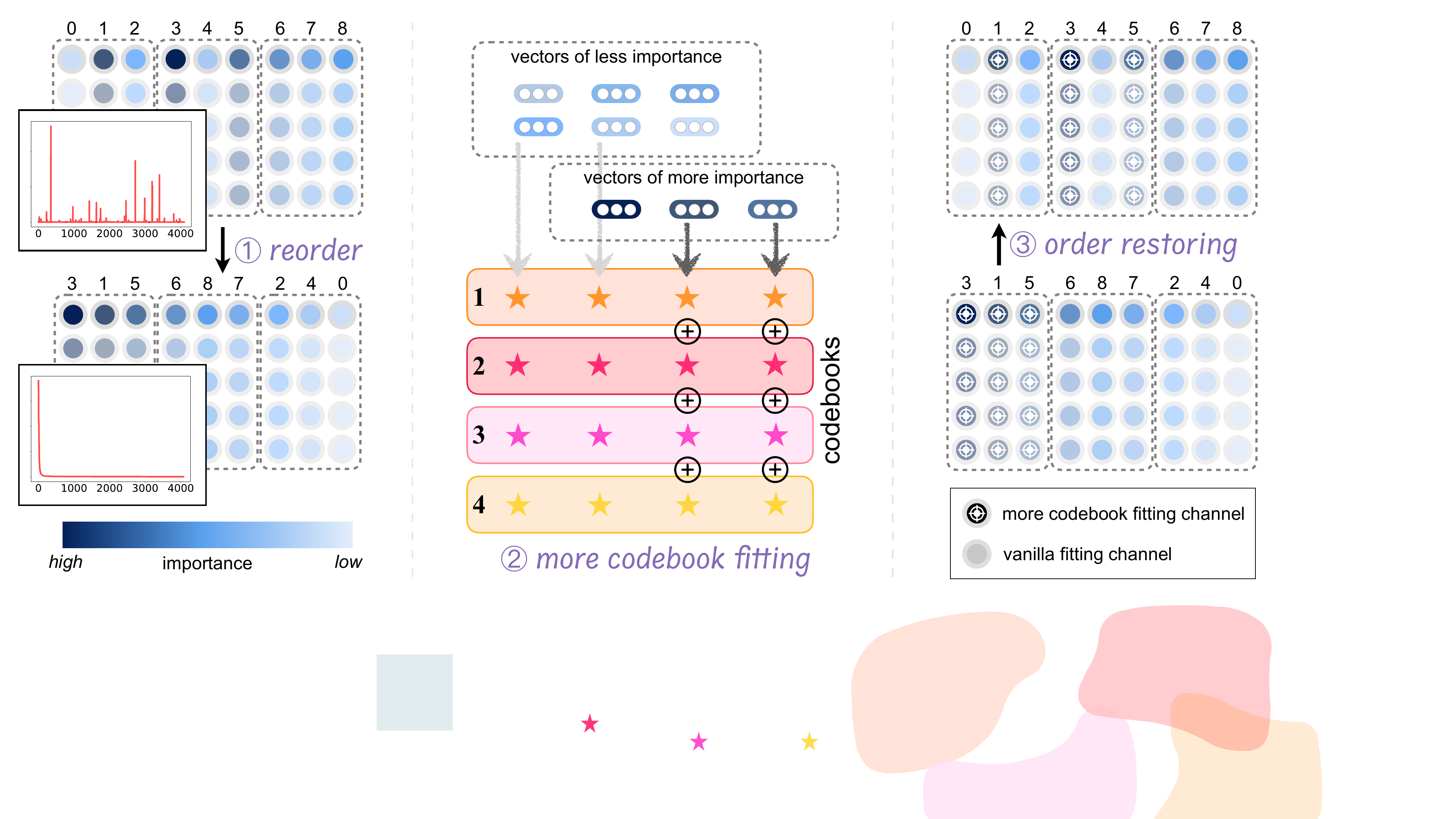}
  \caption{Illustration of the proposed CRVQ, consisting of three phases. On the left, channels are reordered by importance, with circle shading indicating weight significance. Next, vectors are fitted to codebooks, 
  where critical weights are represented as the sum of corresponding vectors from all codebooks (1 to 4).
  The right sub-figure shows channels reordered to their original sequence.}
\label{fig:main}
\end{figure*}

\section{Approach}
\label{sec:approach}

This section shows our design for channel-relaxed quantization. We start with a discussion of how to measure the channel importance in Section~\ref{subsec:measure} and then demonstrate how to adapt different importance into VQ in Section~\ref{subsec:reorder} and~\ref{subsec:fitting}. Finally, we demonstrate the overall algorithm in Section~\ref{subsec:algo}.

\subsection{Measuring Channel Importance}
\label{subsec:measure}

It is widely recognized in recent years that different weight channels exhibit varying levels of sensitivity~\citep{zeroquant2022, smooth2023, spqr2023}. Consequently, the contributions of these channels in the quantization process are not uniform. A considerable number of modern PTQ studies~\citep{spqr2023, llmint2022, smooth2023} leverage this insight by adopting different treatment of weights, leading to improved results. However, the criteria for measuring weight sensitivity vary across different works.

The Hessian metric serves as a widely used benchmark for evaluating weight sensitivity. Inspired by SpQR\citep{spqr2023}, we first employ an importance criterion that jointly considers quantization error and activation magnitude, with the latter derived from the Hessian proxy. Consequently, the importance $I_i$ of a weight channel $\mathbf{W}_{:,i}$ can be formulated as:
\begin{equation}
\label{eq:wa}
    I_i = \max_{j} \frac{\bigl[w_{ji}-\mathrm{VQ}\left(w_{ji}\right)\bigr]^2}{2\left[\mathbf{XX}^\mathrm{T}\right]^{-1}_{ii}},
\end{equation}
where $\mathbf{X}$ denotes as the activation that multiply weight $\mathbf{W}$ to be quantized, see Section~\ref{app:equation} for details. When calculating the importance $I_i$, the quantization error is first obtained by pre-quantizing the weights using VQ and then multiplying activation.

Furthermore, some studies~\citep{zeroquant2022, billm2024} also observe that weights exhibit row-wise salients, leading to varying sensitivity to quantization. Intuitively, the presence of salient weights may pose greater challenges for effective quantization. Thus, channel importance may be defined as $I_i=\max_j w_{ji}^2$ as well. 

The optimal selection in CRVQ will be discussed in Section~\ref{subsec:strategy}. Once the metric $I_i$ for each weight channel is computed, the importance of each channel can be determined.

\subsection{Channel Reordering}
\label{subsec:reorder}

Although the importance of all channels is computed, we are not yet able to handle the more critical channels independently by VQ framework. This is because the distribution of important channels is highly sparse, and their neighboring channels are often not critical. In other words, critical channels are not inherently contiguous. To break this obstacle, we reorder all channels based on their importance. As shown in the left of Figure~\ref{fig:main}, the scattered distribution of important channels (index 1,3 and 5) in the weight matrix is reorganized such that all important channels become adjacent (in the left vector group).

\subsection{Extended Codebook Fitting}
\label{subsec:fitting}

After channel reordering, all channels are grouped column-wise, with each group containing many $d$-dimensional vectors. We define the important channel ratio $\lambda$ and divide all groups into two types, important and non-important. Now, all vectors within a group belong to either important or non-important channels. This grouping allows for special treatment of the important channels.


There are two primary paths to enhance quantization precision, using more codebooks or employing codebooks with larger capacity. However, increasing codebook capacity also increases the encoding bit-width $e$, which in turn inflates the code for non-critical vectors. Therefore, we opt to use additional codebooks instead of expanding the capacity of existing ones. In CRVQ, we designate the codebooks applied to all vector groups as \textbf{basic codebooks} and those specifically used for critical vector groups as \textbf{extended codebooks}. As shown in the middle of Figure~\ref{fig:main}, codebook 1 is the basic codebook, while codebooks 2 to 4 are extended codebooks. We first perform quantization on all groups using the basic codebook.
Next, we supplement the quantization of critical vector groups by fitting them with the extended codebooks using additive VQ. During inference, the weights are restored to their original order through a reverse reordering process. In this paper, we use notation like ``$8+8\times3$'' to denote the codebook configuration, where one basic codebook and three extended codebooks are used.

\subsection{Overall Algorithm}
\label{subsec:algo}

Now we demonstrate our CRVQ with the help of Algorithm~\ref{alg:alg}. Similar to most PTQ methods, the quantization process is performed layer-by-layer. For each weight matrix $\mathbf{W}$ in a specific layer, we first compute the column-wise channel importance and reorder these channels (Line $3\sim5$). Then, we construct the basic codebook $\mathcal{C}_{\mathrm{base}}$ and search the binary encoding $\mathcal{B}_{\mathrm{base}}$ as well as the quantized weight $\mathbf{W}_{\mathrm{encoded}}$ (Line $6\sim7$). Here we use only one basic codebook and please refer to Section~\ref{subsec:aqlm} for more details. Next, we recur to $m-1$ extended codebooks $\{\mathcal{C}^1_\text{ext}, \ldots, \mathcal{C}^{m-1}_\text{ext}\}$ to fitting the $\lambda$ important channels in additive manner (Line $8\sim13$). Additive VQ iteratively \textit{fits the residual} $\mathbf{E}^t$ from the previous quantization step, allowing important vector groups to benefit from multiple rounds of refinement. Finally, we follow previous works~\citep{aqlm2024, quip2024} by finetuning the codebook and beam-searching the optimal code (Line $14\sim17$). Moreover, block and end-to-end finetuning is leveraged for enhancing the quantization performance (Line $19,21$), which is widely used in~\citet{aqlm2024, quip2024}.


\begin{algorithm}[t]
\caption{CRVQ: Channel-Relaxed Vector Quantization for LLMs}
\label{alg:alg}
\small
{\baselineskip=1.15\baselineskip
\begin{algorithmic}[1]
\Require LLM to be quantized $\mathcal{M}$ with blocks $\{B_1, B_2, \ldots, B_k\}$, number of codebooks $m$, ratio of important channels $\lambda$, loss threshold $\epsilon$, and precomputed $\mathbf{X}, \mathbf{XX}^\mathrm{T}$ for each layer.
\For{\texttt{block} $B_i \in \{B_1, B_2, \ldots, B_k\}$}
    \For{\texttt{weight} $\mathbf{W} \in \texttt{layer\_weights}(B_i)$}
        \State $\mathbf{W}_\text{quant} \gets \texttt{Prequantize}(\mathbf{W})$ 
        
        \textcolor{purple!60}{\Comment{prequantize the weight $\mathbf{W}$ using VQ}}
        \State $\mathbf{I} \gets \texttt{ComputeImportance}(\mathbf{W}, \mathbf{W}_\text{quant}, \mathbf{XX}^\textrm{T})$
        \State $\mathbf{W}_\text{sorted} \gets \texttt{ReorderChannels}(\mathbf{W}, \mathbf{I})$
        \State $\mathcal{C}_\text{base} \gets \texttt{KMeansCodebook}(\mathbf{W}_\text{sorted})$
        \State $\mathbf{W}_\text{encoded}, \mathcal{B}_\text{base} \gets \texttt{VectorQuant}(\mathbf{W}_\text{sorted}, \mathcal{C}_\text{base})$ 
        
        \textcolor{purple!60}{\Comment{$\mathcal{B}$ is the encoding of $\mathbf{W}$ on codebook $\mathcal{C}$}}
        \For{$t = 1, \dots, m-1$}
            \State $\mathbf{E}^t \gets \texttt{ComputeError}(\mathbf{W}^\lambda, \mathbf{W}_\text{encoded}^\lambda)$ 
            
            \textcolor{purple!60}{\Comment{$\lambda$ indicates the important channels}}
            \State $\mathcal{C}_\text{ext}^t \gets \texttt{KMeansCodebook}(\mathbf{E}^t)$
            \State $\mathbf{E}^t_\text{encoded}, \mathcal{B}_\text{ext}^t \gets \texttt{VectorQuant}(\mathbf{E}^t, \mathcal{C}_\text{ext}^t)$
            \State $\mathbf{W}_\text{encoded}^\lambda \gets \texttt{Update}(\mathbf{W}_\text{encoded}^\lambda, \mathbf{E}^t_\text{encoded})$
        \EndFor
        \While{$\texttt{QuantLoss}(\mathbf{W}, \mathbf{W}_\text{encoded}, \mathbf{X}) \geq \epsilon$} 
        
        \textcolor{purple!60}{\Comment{loss is $\|\mathbf{WX}-\mathbf{W}_\text{encoded}\mathbf{X}\|_2^2$}}
            \State $\texttt{FineTuneCodebook}(\mathcal{C}_\text{base}, \{\mathcal{C}^1_\text{ext}, \ldots, \mathcal{C}^{m-1}_\text{ext}\})$
            \State $\texttt{BeamSearchOptimize}(\mathbf{W}, \mathcal{C}_\text{base}, $
            \Statex \quad\quad\quad\quad\quad $\{\mathcal{C}^1_\text{ext}, \ldots, \mathcal{C}^{m-1}_\text{ext}\}, \mathcal{B}_\text{base}, \{\mathcal{B}^1_\text{ext}, \ldots, \mathcal{B}^{m-1}_\text{ext}\})$
        \EndWhile
    \EndFor
    \State $\texttt{FineTuneBlock}(B_i)$
\EndFor
\State \textbf{return} $\mathcal{M}_\text{quant} \gets \texttt{E2E\_FineTune}(\mathcal{M})$
\end{algorithmic}
}
\end{algorithm}

\section{Experiment}
\label{sec:experiment}

\subsection{Settings}
\label{subsec:setting}

\paragraph{Models and Data}
We perform experiments on OPT-1.3B/2.7B, LLaMA-7B/13B and LLaMA2-7B/13B to evaluate our method. Results on LLaMA3 and Qwen2 are in Section~\ref{app:models}. Since the baselines used for comparison also require calibration data, we follow the experimental setup of AQLM by randomly sampling fixed-length data from the \texttt{Red\_Pajama} dataset to support algorithm execution. For OPT models, the data length is set to 2048, while for the other models, calibration is performed with a text length of 4096.

\begin{table*}[t!]
\centering
\resizebox{0.95\textwidth}{!}{
\renewcommand{\arraystretch}{1.0}
\begin{tabular}{c@{\hspace{0.20cm}}|@{\hspace{0.40cm}}l@{\hspace{0.40cm}}c@{\hspace{0.40cm}}|c@{\hspace{0.6cm}}c|c@{\hspace{0.30cm}}c@{\hspace{0.30cm}}c@{\hspace{0.30cm}}c@{\hspace{0.30cm}}c@{\hspace{0.30cm}}c@{\hspace{0.30cm}}c}
\toprule
\multirow{2}{*}[-0.8ex]{Model} & \multirow{2}{*}[-0.8ex]{Method} & \multirow{2}{*}[-0.8ex]{Wbits} & \multicolumn{2}{c|}{Perplexity($\downarrow$)} & \multicolumn{7}{c}{Zero-shot Accuracy($\uparrow$)} \\ 
\cmidrule(lr){4-5} \cmidrule(lr){6-12} 
& & & Wiki2 & C4 & Wino. & Hella. & PIQA & BoolQ & ARC-e & ARC-c & Avg. \\ 
\midrule
\noalign{\vskip -1pt}
\multirow{5}{*}[-0ex]{OPT-1.3B} & FP16 & 16.0 & 14.67 & 14.76 & 59.67 & 53.69 & 72.25 & 56.85 & 51.43 & 29.86 & 53.96 \\
\cdashline{2-12}
\noalign{\vskip 3pt}
& PB-LLM & 1.73 & 912.41 & 834.57 & 50.27 & 27.14 & 52.10 & 38.10 & 29.46 & 22.51 & 36.59 \\
& BiLLM & 1.11 & 59.24 & 67.71 & 52.33 & 32.87 & 58.00 & 52.17 & 35.52 & 23.04 & 42.32 \\
& AQLM & 1.02 & 59.68 & 45.82 & 50.75 & 29.94 & 55.71 & \textbf{52.26} & 33.21 & 23.29 & 40.87 \\
& CRVQ & 1.08 & \textbf{39.73} & \textbf{33.29} & \textbf{52.80} & \textbf{32.87} & \textbf{58.43} & 52.11 & \textbf{36.15} & \textbf{23.63} & \textbf{42.67} \\
\noalign{\vskip -1pt}
\midrule
\noalign{\vskip -1pt}
\multirow{5}{*}[-0ex]{OPT-2.7B} & FP16 & 16.0 & 12.46 & 13.17 & 60.69 & 60.64 & 74.59 & 59.60 & 54.34 & 31.23 & 56.85 \\
\cdashline{2-12}
\noalign{\vskip 3pt}
& PB-LLM & 1.73 & 574.13 & 520.86 & 50.27 & 28.07 & 52.80 & 41.14 & 30.41 & 22.51 & 37.53 \\
& BiLLM & 1.11 & 46.40 & 47.95 & 51.85 & 33.82 & 59.41 & \textbf{52.17} & 35.61 & 23.46 & 42.73 \\
& AQLM & 1.01 & 38.73 & 32.51 & 50.43 & 31.61 & 57.56 & 51.77 & 35.61 & 23.81 & 41.80 \\
& CRVQ & 1.08 & \textbf{28.91} & \textbf{26.11} & \textbf{52.88} & \textbf{36.09} & \textbf{60.94} & 51.99 & \textbf{39.35} & \textbf{25.09} & \textbf{44.40} \\
\noalign{\vskip -1pt}
\midrule
\noalign{\vskip -1pt}
\multirow{5}{*}[-0ex]{LLaMA-7B} & FP16 & 16.0 & 5.68 & 7.08 & 69.93 & 76.16 & 79.16 & 74.98 & 72.90 & 44.62 & 69.62 \\
\cdashline{2-12}
\noalign{\vskip 3pt}
& PB-LLM & 1.71 & 231.79 & 272.66 & 50.38 & 28.89 & 53.26 & 44.55 & 31.82 & 23.17 & 38.67 \\
& BiLLM & 1.09 & 45.61 & 75.27 & 52.72 & 35.94 & 58.05 & 56.42 & 35.52 & 22.53 & 43.53 \\
& AQLM & 1.01 & 17.95 & 19.20 & 51.30 & 38.10 & 57.67 & 62.20 & 39.23 & 25.09 & 45.60 \\
& CRVQ & 1.07 & \textbf{13.68} & \textbf{15.48} & \textbf{55.25} & \textbf{43.77} & \textbf{61.48} & \textbf{62.29} & \textbf{44.15} & \textbf{25.43} & \textbf{48.73} \\
\noalign{\vskip -1pt}
\midrule
\noalign{\vskip -1pt}
\multirow{5}{*}[-0ex]{LLaMA-13B} & FP16 & 16.0 & 5.09 & 6.61 & 73.24 & 79.09 & 80.30 & 77.92 & 74.54 & 47.78 & 72.14 \\
\cdashline{2-12}
\noalign{\vskip 3pt}
& PB-LLM & 1.71 & 83.17 & 98.82 & 50.27 & 35.52 & 56.41 & 58.61 & 33.92 & 20.35 & 42.51 \\
& BiLLM & 1.09 & 13.98 & 17.97 & \textbf{61.01} & 51.26 & 67.19 & 62.29 & \textbf{53.11} & 28.61 & \textbf{53.91} \\
& AQLM & 1.01 & 12.93 & 14.67 & 58.25 & 45.61 & 63.11 & 63.43 & 46.51 & 25.17 & 50.35 \\
& CRVQ & 1.06 & \textbf{10.73} & \textbf{12.56} & 60.22 & \textbf{52.04} & \textbf{67.25} & \textbf{64.37} & 50.76 & \textbf{28.75} & 53.90 \\
\noalign{\vskip -1pt}
\midrule
\noalign{\vskip -1pt}
\multirow{5}{*}[-0ex]{LLaMA2-7B} & FP16 & 16.0 & 5.12 & 6.63 & 69.22 & 75.93 & 79.11 & 77.68 & 74.75 & 46.16 & 70.47 \\
\cdashline{2-12}
\noalign{\vskip 3pt}
& PB-LLM & 1.71 & 186.06 & 223.54 & 49.48 & 31.10 & 52.64 & 40.82 & 30.41 & 24.20 & 38.11 \\
& BiLLM & 1.08 & 26.52 & 39.71 & 50.59 & 31.46 & 55.01 & 60.10 & 35.16 & 20.44 & 42.12 \\
& AQLM & 1.01 & 21.28 & 24.44 & 50.59 & 32.04 & 55.66 & 61.41 & 33.00 & 20.73 & 42.24 \\
& CRVQ & 1.07 & \textbf{13.01} & \textbf{15.72} & \textbf{55.09} & \textbf{42.28} & \textbf{61.48} & \textbf{62.57} & \textbf{39.31} & \textbf{25.51} & \textbf{47.71} \\
\noalign{\vskip -1pt}
\midrule
\noalign{\vskip -1pt}
\multirow{5}{*}[-0ex]{LLaMA2-13B} & FP16 & 16.0 & 4.57 & 6.05 & 71.90 & 79.37 & 80.63 & 80.86 & 77.53 & 49.23 & 73.25 \\
\cdashline{2-12}
\noalign{\vskip 3pt}
& PB-LLM & 1.71 & 228.22 & 259.69 & 50.56 & 30.82 & 53.98 & 42.18 & 29.94 & 22.80 & 38.38 \\
& BiLLM & 1.08 & 20.52 & 32.01 & 54.85 & 38.95 & 60.94 & 65.60 & 45.96 & 25.94 & 48.71 \\
& AQLM & 1.01 & 15.25 & 18.35 & 52.49 & 35.14 & 56.42 & 62.29 & 38.01 & 24.32 & 44.78 \\
& CRVQ & 1.06 & \textbf{9.81} & \textbf{12.48} & \textbf{58.09} & \textbf{50.88} & \textbf{65.40} & \textbf{67.89} & \textbf{49.49} & \textbf{28.07} & \textbf{53.30} \\
\noalign{\vskip -1pt}
\bottomrule
\end{tabular}
}
\vspace{-0.3em}
\caption{Main results of evaluation experiment. We report the perplexity and zero-shot accuracy. The results of AQLM and CRVQ are based on the models after e2e-finetuning. The \textbf{best} scores are highlighted in bold.}
\label{tab:main}
\vspace{-0.5em}
\end{table*}

\paragraph{Baselines}
We compare CRVQ with several representative extreme PTQ methods, including PB-LLM~\citep{pbllm2024}, BiLLM~\citep{billm2024}, and AQLM~\citep{aqlm2024}. PB-LLM is integrated with GPTQ as its backbone and 10\% of critical weights quantized to 8 bits to ensure an average bit-width below 2 bits. In BiLLM, we use a block size of 128 and 128 calibration samples. In AQLM, we configure the codebook count $m = 1$, vector dimension $d = 8$, and codebook bit-width $e = 8$. We adopt its optimal setup by using 2048 calibration samples. For our proposed CRVQ, the ratio of critical channels is set to $\lambda = 2\%$, with \textbf{one} basic codebook and \textbf{three} extended codebooks. Similar to AQLM, CRVQ employs $d = 8$, $e = 8$, and 2048 calibration samples. All experiments are conducted on 1$\times$NVIDIA A100-80GB. Further details on the baselines are provided in Section~\ref{app:details}.

\paragraph{Evaluation Metrics}
To evaluate baseline performance, we calculate perplexity on datasets like WikiText2~\citep{wiki22016} and C4~\citep{C42020}, using randomly sampled evaluation data of the same length as the calibration data. Lower perplexity indicates better preservation of the original output distribution. Additionally, we report zero-shot accuracies on commonsense reasoning tasks, including Winogrande~\citep{winogrande2021}, HellaSwag~\citep{hellaswag2019}, PIQA~\citep{piqa2020}, BoolQ~\citep{boolq2019}, and ARC-e/ARC-c~\citep{arc2018}. Evaluations are conducted using the open-source toolkit LM-Evaluation-Harness\footnote{\url{https://github.com/EleutherAI/lm-evaluation-harness}}.

\begin{figure*}[t!]
    \centering
    \begin{subfigure}[b]{0.32\textwidth}
        \includegraphics[width=1.00\textwidth]{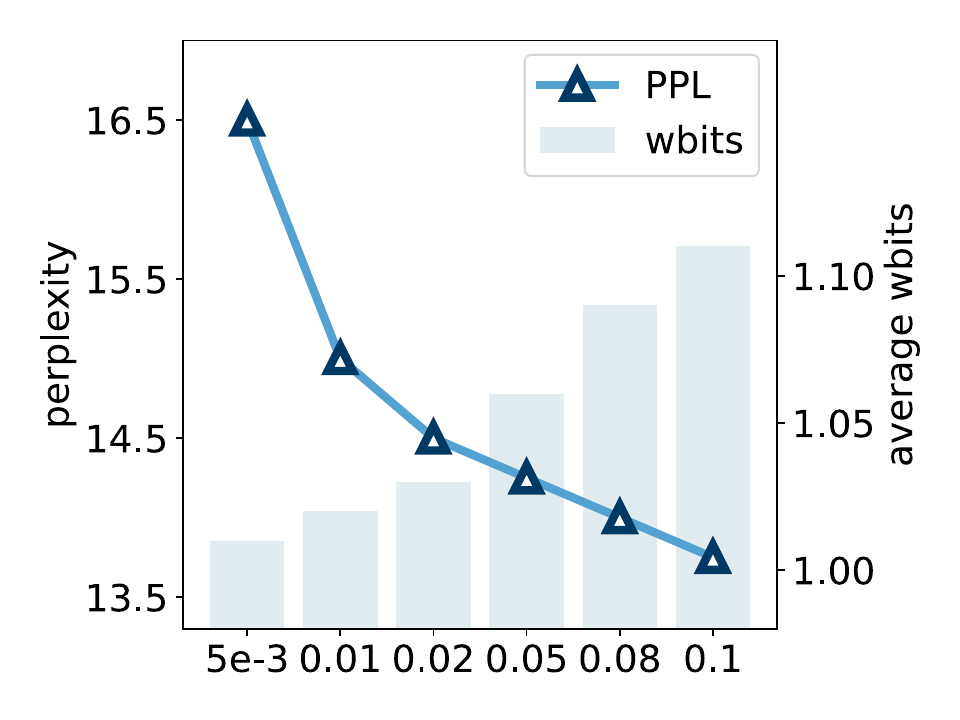}
        \caption{different $\lambda$}
        \label{fig:lambda}
    \end{subfigure}
    \begin{subfigure}[b]{0.32\textwidth}
        \includegraphics[width=1.00\textwidth]{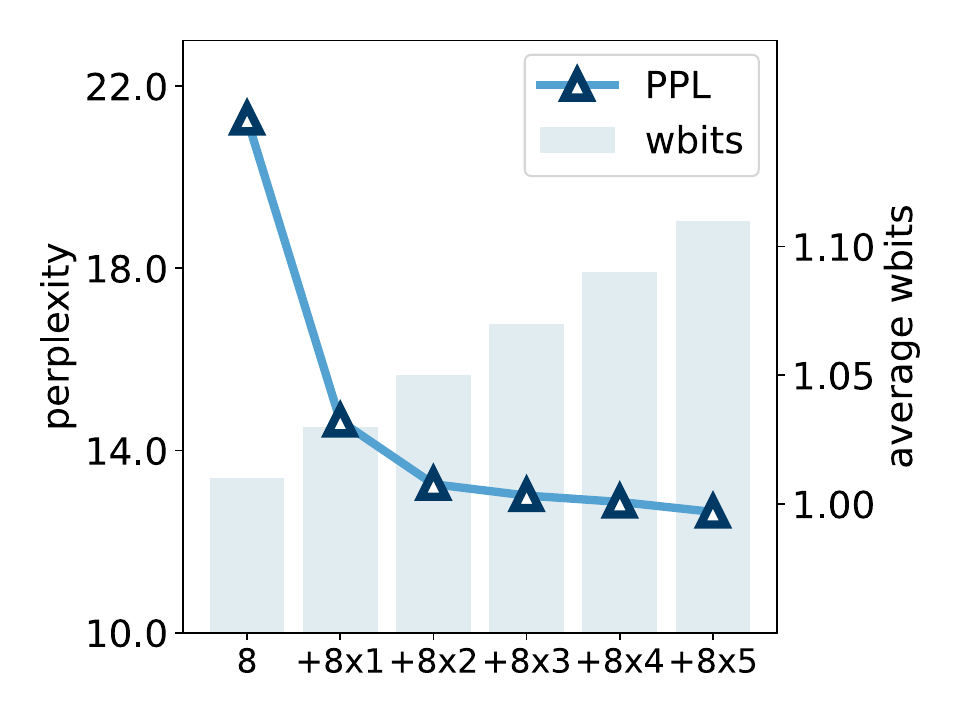}
        \caption{different $m$ ($\lambda=0.02$)}
        \label{fig:0.02}
    \end{subfigure}
    \begin{subfigure}[b]{0.32\textwidth}
        \includegraphics[width=1.00\textwidth]{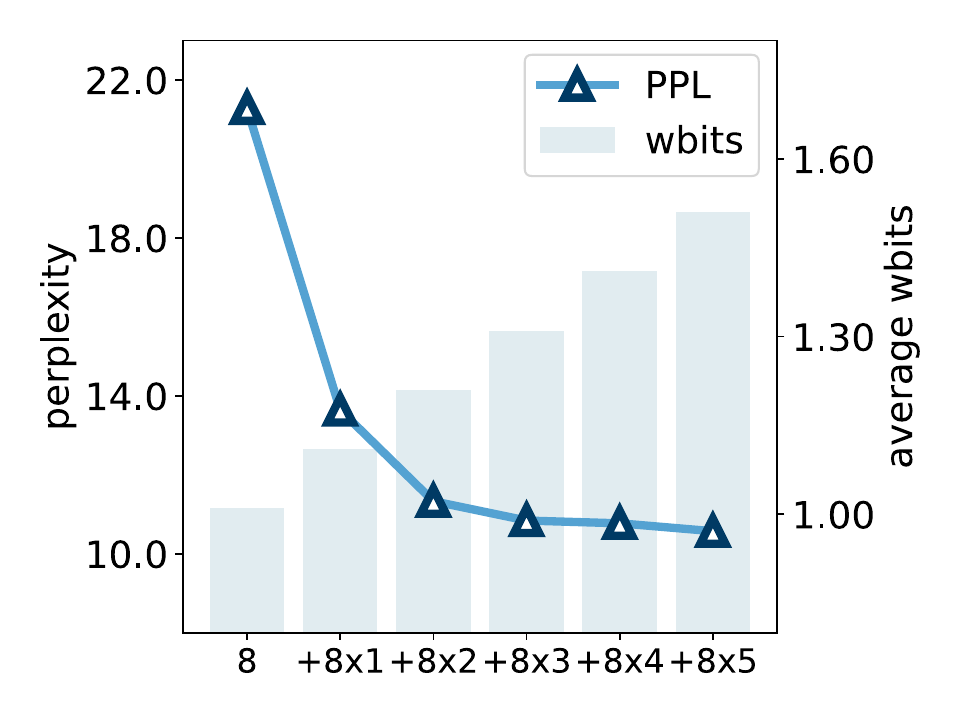}
        \caption{different $m$ ($\lambda=0.1$)}
        \label{fig:0.1}
    \end{subfigure}
    \vspace{-0.5em}
    \caption{Quantization performance varies with the critical channel ratio $\lambda$ and codebook count $m$.}
    \label{fig:m}
    \vspace{-1.0em}
\end{figure*}

\subsection{Main Results}
\label{subsec:result}

As shown in Table~\ref{tab:main}, CRVQ consistently outperforms baselines across model sizes. While PB-LLM has the highest average bit-width, it shows no clear advantage in perplexity or zero-shot tasks, likely due to suboptimal bit utilization. In contrast, BiLLM, AQLM, and CRVQ perform comparably. Extreme PTQ incurs some performance loss compared to QAT~\citep{onebit2024, bmos2024}, but its benefits scale with model size. Smaller models like OPT show performance degradation, whereas larger LLaMA models achieve results close to QAT~\citep{onebit2024} or even unquantized models~\citep{llama2023se}.

For perplexity, CRVQ achieves the lowest across all models, with AQLM as the second-best. For example, on WikiText2 with LLaMA-13B, CRVQ achieves a perplexity of 9.81, compared to 15.25 for AQLM, demonstrating their effectiveness in reducing bit-width. More importantly, CRVQ achieves this with negligible additional bit-width cost.

For zero-shot accuracy, CRVQ achieves the best results in most cases, particularly within the LLaMA series, except for a few tasks on LLaMA-13B (e.g., Winogrande, ARC-e). On most LLaMA models, CRVQ outperforms the second-best baseline by 7\% to 13\%.

\section{Discussion}
\label{sec:disscusion}

This section delves into the effectiveness and its inner components of the proposed method on LLaMA2-7B, with the goal of enhancing understanding of its underlying mechanisms.

\subsection{Ratio of Important Channels}
\label{subsec:ratio}

CRVQ leverages extra codebooks for a few critical channels to maximize quantization benefits with minimal overhead. 
A natural question arises: \textit{how many channels are sufficient for CRVQ to perform effectively?} We test this under codebook count $m=2$, i.e. ``$8+8\times1$'', varying the channel ratio $\lambda$ from 5e-3 to 0.1. As shown in Figure~\ref{fig:lambda}, when $\lambda < 0.02$, the perplexity of the quantized model drops rapidly by 6.7 compared to the model without critical channels. Increasing $\lambda$ to 0.1 further improves performance, even though the average bit-width remains as low as 1.1 bit. These results validate the significant impact of critical channels. Considering the performance-to-cost ratio at $\lambda = 0.02$, this setting maybe well-suited for deployment on extremely low-resource devices.

\subsection{Number of Extended Codebooks}
\label{subsec:number}

In CRVQ, we apply multiple extended codebooks to a few critical channels. There are also two important questions: \textit{how many extended codebooks are necessary for the method to work effectively, and is more always better?} We investigate the impact of the extended codebooks under two critical channel ratios, $\lambda = 0.02$ and $\lambda = 0.1$. As shown clearly in Figure~\ref{fig:0.02} and~\ref{fig:0.1}, using three extended codebooks provides the most notably performance improvement compared to using none in both settings. However, when the extended codebooks count exceeds three, the performance gains become marginal. Thus, adding more codebooks does not necessarily yield better results and we recommend using three extended codebooks for optimal performance.

\begin{table*}[t!]
\centering
\resizebox{0.80\textwidth}{!}{
\renewcommand{\arraystretch}{1.0}
\begin{tabular}{c@{\hspace{0.2cm}}|l@{\hspace{0.20cm}}|c@{\hspace{0.15cm}}c@{\hspace{0.15cm}}|c@{\hspace{0.25cm}}c@{\hspace{0.25cm}}c@{\hspace{0.25cm}}c@{\hspace{0.25cm}}c@{\hspace{0.25cm}}c@{\hspace{0.25cm}}c@{\hspace{0.25cm}}}
\toprule
\multirow{2}{*}[-0.6ex]{Ratio} & \multirow{2}{*}[-0.6ex]{\begin{tabular}[c]{@{}c@{}}Reorder\\Strategy\end{tabular}} & \multicolumn{2}{c|}{Perplexity($\downarrow$)} & \multicolumn{7}{c}{Zero-shot Accuracy($\uparrow$)} \\ 
\cmidrule(lr){3-4} \cmidrule(lr){5-11} 
& & Wiki2 & C4 & Wino. & Hella. & PIQA & BoolQ & ARC-e & ARC-c & Avg. \\ 
\midrule
\noalign{\vskip -1pt}
\multirow{3}{*}[-0ex]{$\lambda =0.02$} & \textit{Random} & 19.99 & 22.93 & 50.12 & 32.87 & 55.44 & 61.25 & 34.09 & 22.70 & 42.75 \\
& \textit{W-only} & 17.72 & 20.19 & 53.51 & 34.69 & 56.04 & 57.43 & 36.95 & 24.74 & 43.89 \\
& \textit{W-A} & \textbf{13.01} & \textbf{15.72} & \textbf{55.09} & \textbf{42.28} & \textbf{61.48} & \textbf{62.57} & \textbf{39.31} & \textbf{25.51} & \textbf{47.71} \\
\noalign{\vskip -1pt}
\midrule
\noalign{\vskip -1pt}
\multirow{3}{*}[-0ex]{$\lambda =0.1$} & \textit{Random} & 14.89 & 17.88 & 53.83 & 37.73 & 58.00 & 57.09 & 37.21 & 23.98 & 44.64 \\
& \textit{W-only} & 16.32 & 18.24 & 51.22 & 37.66 & 58.81 & 59.02 & 38.30 & 23.98 & 44.83 \\
& \textit{W-A} & \textbf{10.83} & \textbf{13.43} & \textbf{56.99} & \textbf{46.95} & \textbf{64.64} & \textbf{61.59} & \textbf{44.61} & \textbf{26.71} & \textbf{50.25} \\
\noalign{\vskip -1pt}
\bottomrule
\end{tabular}
}
\vspace{-0.3em}
\caption{Performance on different reorderding strategy. Here we use $8+8\times3$ setting.}
\label{tab:shuffle}
\vspace{-0.5em}
\end{table*}

\subsection{Different Reorder Strategy}
\label{subsec:strategy}

We discuss on different importance metrics to understand the contributions of different channels in CRVQ. Based on the analysis in Section~\ref{subsec:measure}, we evaluate three channel reordering strategies: random reordering (\textit{Random}), reordering based on weight magnitude (\textit{W-only}), and reordering using weight combined with Hessian proxy (\textit{W-A}). The results in Table~\ref{tab:shuffle} indicate that the \textit{W-A} consistently outperforms the others. We also find that the effectiveness of \textit{Random} and \textit{W-only} strategies varies with $\lambda$. Specifically, \textit{W-only} performs better than \textit{Random} at extremely low $\lambda$. It is likely because the \textit{Random} strategy has low hit rates for critical channels under such condition. These results emphasize the necessity of explicitly catching critical channels in VQ to achieve superior performance.

\begin{figure}[t]
  \centering
  \includegraphics[width=\columnwidth]{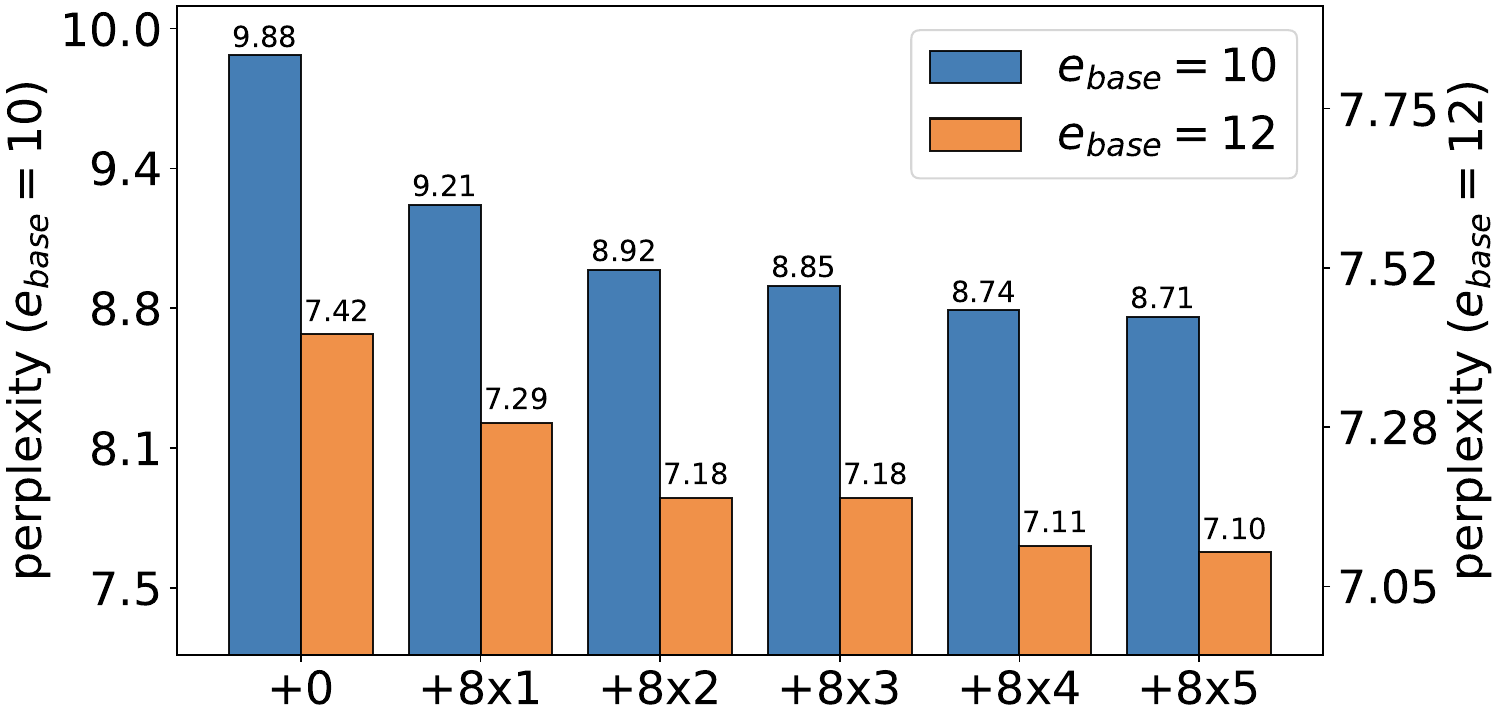}
  \caption{Effectiveness of CRVQ when combine with stronger basic codebook, i.e. $e$ changes from 8 to 10 and 12. The performance consistently improves as the extended codebook count increases.}
\label{fig:c}
\vspace{-1em}
\end{figure}

\subsection{Flexible Codebook Settings}
\label{subsec:combine}

CRVQ is not limited to integrate with a single basic codebook and $e=8$. On the contrary, it can be seamlessly integrated with any vanilla VQ method, applying multi-codebook fitting specially to critical channels. First, as the basic codebook bit-width increases, its fitting capability improves, but CRVQ still works well. Figure~\ref{fig:c} presents the performance of CRVQ when applied to two larger codebooks ($e$ = 10 and $e$ = 12). Obviously, CRVQ consistently shows effectiveness as the extended codebooks increase. Moreover, our method also works well with involving more base codebooks. For instance, under the setting of $m = 2, d = 16, e = 8$, the perplexity of AQLM is 17.11, whereas CRVQ achieves 14.21 with three extended codebooks. This demonstrates the generality and flexibility of our approach across diverse settings.

\subsection{Hardware-friendly Deployment}
\label{subsec:deployment}

Although traditional VQ can also enhance its performance by increasing $e$, we recommend using $e = 8$ in combination with additional extended codebooks when deploying models. First, increasing $e$ (e.g., to $e = 10$) may result in a larger overall bit-width, leading to lower bit-width utilization. It also hampers hardware alignment and computational acceleration. Furthermore, combining hardware-friendly basic codebooks with several extended codebooks and adjusting $\lambda \%$ critical channels can efficiently achieve better quantization performance without these drawbacks. Therefore, CRVQ is cost-effective on low-resource devices and easily adaptable to different devices.

\subsection{Comparison to QAT method}
\label{subsec:qat}

Our work focuses on extreme algorithms within the PTQ area. A related direction with similar goal is QAT, and the two methods have their respective advantages and limitations. QAT achieves relatively better performance through extensive training, whereas PTQ avoids the long training process but often falls short of QAT in performance. We compare our method with the classical 1-bit QAT approach, OneBit~\citep{onebit2024}, as shown in Table~\ref{tab:onebit}. Notably, our algorithm outperforms OneBit in terms of computational efficiency while achieving comparable performance. Our study shows that with continued advancements, PTQ is approaching the performance levels of QAT.

\begin{table}[t]
\centering
\resizebox{\columnwidth}{!}{
\renewcommand{\arraystretch}{1.0}
\begin{tabular}{l|cccc}
\toprule
Method & Wiki2 & \#GPUs & \#Hours /GPU & \#Hours \\ 
\midrule
OneBit & 8.76 & 16 & 448 & 7168 \\
CRVQ & 9.81 & 1 & 50 & 50 \\
\bottomrule
\end{tabular}
}
\vspace{-0.3em}
\caption{Comparison between OneBit and CRVQ on LLaMA2-13B. Here we use $8+8\times3$ setting.}
\label{tab:onebit}
\vspace{-1em}
\end{table}

\section{Conclusion}
\label{sec:conclusion}

We introduce CRVQ, an simple yet effective extreme quantization algorithm that enhances performance by identifying critical weight channels and applying multi-codebook fitting via channel reordering. Extensive experiments on models of varying sizes demonstrate significant performance gains at minimal additional cost, outperforming representative strong baselines. We also analyze key factors affecting CRVQ and provide guidance for hardware resource adaptation.

\section*{Limitations}

While our method achieves obvious post-training extreme compression with minimal cost, offering a cost-effective solution for deploying LLMs on low-resource devices, some limitations remain. 
First, like other 1-bit baselines~\citep{onebit2024, billm2024}, our approach exhibits performance degradation compared to the original model and higher-bit quantization methods (e.g., 2-bit and above), highlighting the need for further exploration into the limits of extreme compression. 
Additionally, although our method significantly reduces computational demands compared to QAT-based 1-bit quantization~\citep{onebit2024}, the quantization process itself remains time-consuming. Fortunately, this is a one-time cost. 
Finally, similar to many classic quantization studies, the role of a small subset of critical channels in the quantization process and its impact on overall performance remain insufficiently explored. This opens up exciting avenues for future research.

\section*{Ethics Statement}

In this study, we employ models that are publicly available and open source. We affirm that the use of these models aligns with their original intended purposes. These models have been utilized strictly within the scope of academic and research-based activities, adhering to ethical guidelines and ensuring compliance with open-source licenses.



\bibliography{custom}

\appendix

\section{Appendix}
\label{sec:appendix}

\subsection{Average Bits Calculation}
\label{app:bits}

We first compute the average bit-width of quantized weights using the classical vector quantization method and explain the derivation of Equation~\ref{eq:bitwidth}, with a 4096$\times$4096 matrix as an example. Following the notations in this paper, let $m$ denote the number of codebooks, $e$ the codebook bit-width, $d$ the dimension of the short vectors obtained from matrix partitioning, and $n$ the average bit-width. The total space consumption consists of two parts: the codebooks and the binary encodings.

Consider the total number of vectors. Each row is divided into $2^{12}/d$ vectors, and the entire matrix is partitioned into $2^{24}/d$ vectors. If each vector is represented using $e$-bit binary encoding, the total code space required is $2^{24}e/d$ bits. Since each codebook has its own code, the total code overhead is $2^{24}me/d$ bits. 
Additionally, each codebook contains $2^e$ vectors of length $d$, where each value is stored in FP16 format. Therefore, the total space occupied by $m$ codebooks is $2^emd\times2^4$ bits. Thus, the total storage requirement is given by $2^{24}me/d + 2^{e+4}md$. Since the matrix contains $2^{24}$ elements, the quantized average bit-width is:
\begin{equation}
    n = \frac{me}{d} + 2^{e-20}md. \notag
\end{equation}
Assuming that the setting of AQLM is $m=1, e=8, d=8$, the average bit-width here is $n\approx1.002$.

CRVQ introduces the concept of extended codebooks. Suppose that among the $m$ codebooks, there are $m-1$ extended codebooks used to encode $\lambda$-fraction of important vectors. The total code space for these important vectors can be expressed as $(m-1)\lambda e/d$. Moreover, storing column reordering indices requires extra space. Each index occupies 16 bits (with 12 effective bits), resulting in an additional storage requirement of $2^{12} \times 2^4 / 2^{24}$  bits. Clearly, this overhead is negligible. Thus, the total storage requirement for CRVQ is given by:
\begin{equation}
    n = \frac{e}{d} + \frac{(m-1)\lambda e}{d} + 2^{e-20}md.
\end{equation}
Assuming that the setting of CRVQ is $8+8\times3$ (i.e., $m=4$), $d=8$ and $\lambda=0.02$, the average bit-width here is $n\approx1.062$.

\subsection{Hessian-based Importance Metric}
\label{app:equation}

As shown in Equation~\ref{eq:wa}, the importance metric based on the Hessian proxy considers both weight quantization error and activation magnitude. To understand the computation of importance, we unpack the matrix formulation. Let $\mathbf{W}^{M\times N}$ denote the weight matrix to be quantized, and the quantization error for each weight can be represented by $\Delta w_{ji}=w_{ji}-\mathrm{VQ}(w_{ji})$. Suppose the activation multiplying with the weight is denoted as $\mathbf{X}^{N\times O}$. Then, the Hessian proxy can be computed as:
\begin{equation}
    \mathbf{XX}^\mathrm{T}=\begin{pmatrix}
 \sum_{k=1}^{O}x_{1k}^2 &  \cdots  & \sum_{k=1}^{O}x_{1k}x_{Nk} \\
 \vdots & \ddots & \vdots  \\
 \sum_{k=1}^{O}x_{Nk}x_{1k} & \cdots & \sum_{k=1}^{O}x_{Nk}^2
\end{pmatrix}
\end{equation}
Hence the Hessian-based importance metric can be reformulated as:
\begin{equation}
    I_i = \max_{j}\frac{1}{2}\sum_{k=1}^{O}(x_{ik}\Delta w_{ji})^2.
\end{equation}
Therefore, the selection of critical channels is guided by the outcome of the matrix product. Channels with both high quantization error and large corresponding activation magnitude have the greatest impact on the quantization results.

\subsection{Results on LLaMA3 and Qwen2}
\label{app:models}

To further validate the effectiveness of CRVQ across more LLMs, we extend our experiments to LLaMA3 and Qwen2 models, ranging in size from 0.5B to 8B. Except for the differences in model series, all experimental settings remain consistent with those in Section~\ref{subsec:setting}, including the use of \texttt{Red\_Pajama} as the calibration dataset. Additionally, the evaluation datasets follow the same configuration as described in Section~\ref{subsec:setting}. 

Here, we compare CRVQ against two strong baselines, BiLLM and AQLM, with results presented in Table~\ref{tab:models}. 
The experiments show that, in terms of perplexity, CRVQ consistently outperforms all baselines across different models. For zero-shot accuracy, CRVQ surpasses other baselines on most datasets, achieving overall superior performance.

\begin{table*}[t!]
\centering
\resizebox{0.95\textwidth}{!}{
\renewcommand{\arraystretch}{1.0}
\begin{tabular}{c@{\hspace{0.20cm}}|@{\hspace{0.40cm}}l@{\hspace{0.40cm}}c@{\hspace{0.40cm}}|c@{\hspace{0.6cm}}c|c@{\hspace{0.30cm}}c@{\hspace{0.30cm}}c@{\hspace{0.30cm}}c@{\hspace{0.30cm}}c@{\hspace{0.30cm}}c@{\hspace{0.30cm}}c}
\toprule
\multirow{2}{*}[-0.8ex]{Model} & \multirow{2}{*}[-0.8ex]{Method} & \multirow{2}{*}[-0.8ex]{Wbits} & \multicolumn{2}{c|}{Perplexity($\downarrow$)} & \multicolumn{7}{c}{Zero-shot Accuracy($\uparrow$)} \\ 
\cmidrule(lr){4-5} \cmidrule(lr){6-12} 
& & & Wiki2 & C4 & Wino. & Hella. & PIQA & BoolQ & ARC-e & ARC-c & Avg. \\ 
\midrule
\noalign{\vskip -1pt}
\multirow{5}{*}[1ex]{LLaMA3.2-1B} & FP16 & 16.0 & 9.07 & 12.04 & 60.14 & 63.62 & 74.59 & 63.98 & 60.52 & 36.01 & 59.81 \\
\cdashline{2-12}
\noalign{\vskip 3pt}
& BiLLM & 1.11 & 394.87 & 398.45 & \textbf{50.43} & \textbf{28.72} & 53.26 & 58.13 & 29.50 & \textbf{23.89} & 40.65 \\
& AQLM & 1.01 & 115.25 & 90.85 & 47.91 & 26.95 & 53.92 & 50.00 & 30.85 & 20.65 & 38.38 \\
& CRVQ & 1.06 & \textbf{70.03} & \textbf{64.72} & 48.70 & 28.15 & \textbf{54.95} & \textbf{61.47} & \textbf{31.19} & 21.59 & \textbf{41.01} \\
\noalign{\vskip -1pt}
\midrule
\noalign{\vskip -1pt}
\multirow{5}{*}[1ex]{LLaMA3.2-3B} & FP16 & 16.0 & 7.28 & 10.01 & 69.38 & 73.61 & 77.42 & 72.94 & 71.59 & 45.90 & 68.47 \\
\cdashline{2-12}
\noalign{\vskip 3pt}
& BiLLM & 1.11 & 122.19 & 139.83 & \textbf{51.14} & \textbf{32.01} & 54.90 & 61.80 & \textbf{32.87} & 21.25 & 42.31 \\
& AQLM & 1.01 & 83.03 & 63.99 & 50.04 & 28.16 & 53.97 & 61.62 & 30.89 & 21.16 & 40.97 \\
& CRVQ & 1.08 & \textbf{49.17} & \textbf{40.23} & 50.91 & 30.98 & \textbf{56.04} & \textbf{62.45} & 32.03 & \textbf{21.42} & \textbf{42.31} \\
\noalign{\vskip -1pt}
\midrule
\noalign{\vskip -1pt}
\multirow{5}{*}[1ex]{LLaMA3.1-8B} & FP16 & 16.0 & 5.84 & 8.43 & 73.48 & 78.85 & 81.18 & 82.02 & 81.19 & 53.50 & 75.03 \\
\cdashline{2-12}
\noalign{\vskip 3pt}
& BiLLM & 1.09 & 38.79 & 57.07 & 50.43 & \textbf{36.16} & 57.94 & \textbf{62.57} & 35.19 & 23.04 & 44.22 \\
& AQLM & 1.01 & 57.90 & 48.44 & 49.57 & 28.77 & 54.19 & 57.71 & 30.13 & 20.22 & 40.09 \\
& CRVQ & 1.07 & \textbf{27.36} & \textbf{29.53} & \textbf{51.62} & 36.10 & \textbf{59.92} & 61.53 & \textbf{37.25} & \textbf{23.29} & \textbf{44.95} \\
\noalign{\vskip -1pt}
\midrule
\noalign{\vskip -1pt}
\multirow{5}{*}[1ex]{Qwen2-0.5B} & FP16 & 16.0 & 12.35 & 16.18 & 57.46 & 49.07 & 69.31 & 60.86 & 50.25 & 28.67 & 52.60 \\
\cdashline{2-12}
\noalign{\vskip 3pt}
& BiLLM & 1.09 & 1549.81 & 1587.81 & \textbf{51.54} & 26.79 & 51.90 & 45.32 & 28.03 & \textbf{23.21} & 37.79 \\
& AQLM & 1.01 & 96.34 & 84.86 & 48.54 & \textbf{27.26} & 52.18 & 54.68 & 31.06 & 21.33 & 39.17 \\
& CRVQ & 1.06 & \textbf{77.72} & \textbf{69.06} & 50.59 & 26.87 & \textbf{54.46} & \textbf{61.28} & \textbf{31.36} & 22.78 & \textbf{41.22} \\
\noalign{\vskip -1pt}
\midrule
\noalign{\vskip -1pt}
\multirow{5}{*}[1ex]{Qwen2-1.5B} & FP16 & 16.0 & 8.87 & 12.20 & 66.38 & 65.37 & 75.35 & 72.42 & 60.77 & 36.26 & 62.75 \\
\cdashline{2-12}
\noalign{\vskip 3pt}
& BiLLM & 1.08 & 522.46 & 618.99 & 50.36 & 28.45 & 53.37 & 52.81 & 28.45 & \textbf{23.72} & 39.52 \\
& AQLM & 1.01 & 52.24 & 51.97 & \textbf{53.35} & 29.10 & 56.26 & 62.05 & 31.02 & 21.59 & 42.22 \\
& CRVQ & 1.07 & \textbf{44.62} & \textbf{43.66} & 51.70 & \textbf{29.68} & \textbf{56.47} & \textbf{62.23} & \textbf{31.40} & 22.18 & \textbf{42.28} \\
\noalign{\vskip -1pt}
\midrule
\noalign{\vskip -1pt}
\multirow{5}{*}[1ex]{Qwen2-7B} & FP16 & 16.0 & 6.67 & 9.57 & 72.45 & 78.85 & 81.01 & 84.98 & 74.49 & 49.91 & 73.61 \\
\cdashline{2-12}
\noalign{\vskip 3pt}
& BiLLM & 1.08 & 28.80 & 31.68 & 56.27 & 46.01 & 64.85 & 62.81 & \textbf{49.54} & \textbf{28.16} & 51.27 \\
& AQLM & 1.01 & 20.57 & 23.03 & 52.88 & 42.30 & 62.46 & 62.02 & 41.62 & 25.34 & 47.77 \\
& CRVQ & 1.06 & \textbf{17.39} & \textbf{19.91} & \textbf{57.54} & \textbf{47.18} & \textbf{65.13} & \textbf{64.04} & 46.25 & 27.56 & \textbf{51.28} \\
\noalign{\vskip -1pt}
\bottomrule
\end{tabular}
}
\caption{Perplexity and zero-shot accuracy on LLaMA3 on Qwen2 models. The results of AQLM and CRVQ are based on the models after e2e-finetuning. The \textbf{best} scores are highlighted in bold.}
\label{tab:models}
\end{table*}

\begin{table*}[t!]
\centering
\resizebox{0.95\textwidth}{!}{
\renewcommand{\arraystretch}{1.0}
\begin{tabular}{c@{\hspace{0.20cm}}|@{\hspace{0.40cm}}l@{\hspace{0.40cm}}c@{\hspace{0.40cm}}|c@{\hspace{0.6cm}}c|c@{\hspace{0.30cm}}c@{\hspace{0.30cm}}c@{\hspace{0.30cm}}c@{\hspace{0.30cm}}c@{\hspace{0.30cm}}c@{\hspace{0.30cm}}c}
\toprule
\multirow{2}{*}[-0.8ex]{Model} & \multirow{2}{*}[-0.8ex]{Method} & \multirow{2}{*}[-0.8ex]{Wbits} & \multicolumn{2}{c|}{Perplexity($\downarrow$)} & \multicolumn{7}{c}{Zero-shot Accuracy($\uparrow$)} \\ 
\cmidrule(lr){4-5} \cmidrule(lr){6-12} 
& & & Wiki2 & C4 & Wino. & Hella. & PIQA & BoolQ & ARC-e & ARC-c & Avg. \\ 
\midrule
\noalign{\vskip -1pt}
\multirow{5}{*}[2.7ex]{Mistral-7B} & FP16 & 16.0 & 4.91 & 7.42 & 74.11 & 81.11 & 82.15 & 83.76 & 79.59 & 54.18 & 75.81 \\
\cdashline{2-12}
\noalign{\vskip 3pt}
& AQLM & 1.01 & 16.28 & 19.84 & 52.25 & 36.33 & 57.62 & 61.74 & 37.88 & 23.81 & 44.94 \\
& CRVQ & 1.06 & \textbf{13.01} & \textbf{16.23} & \textbf{54.70} & \textbf{44.11} & \textbf{60.83} & \textbf{62.75} & \textbf{42.26} & \textbf{25.51} & \textbf{48.36} \\
\noalign{\vskip -1pt}
\midrule
\noalign{\vskip -1pt}
\multirow{5}{*}[2.7ex]{Mistral-8$\times$7B} & FP16 & 16.0 & 3.59 & 6.52 & 76.64 & 83.94 & 83.35 & 85.41 & 83.42 & 59.81 & 78.76 \\
\cdashline{2-12}
\noalign{\vskip 3pt}
& AQLM & 1.01 & 12.42 & 15.93 & 57.06 & 48.86 & 66.87 & 63.49 & 49.71 & 26.88 & 52.14 \\
& CRVQ & 1.08 & \textbf{9.28} & \textbf{12.32} & \textbf{60.69} & \textbf{63.60} & \textbf{73.56} & \textbf{72.32} & \textbf{57.53} & \textbf{34.47} & \textbf{60.36} \\
\noalign{\vskip -1pt}
\bottomrule
\end{tabular}
}
\caption{Perplexity and zero-shot accuracy on Mistral-7B / 8$\times$7B models. The results of AQLM and CRVQ are based on the models after e2e-finetuning. The \textbf{best} scores are highlighted in bold.}
\vspace{-0.8em}
\label{tab:mistral}
\end{table*}

\subsection{Results on Mistral-7B / 8$\times$7B}
\label{app:mistral}

We also extend the application of CRVQ to the Mistral model\footnote{\url{https://mistral.ai/en/models}}. Furthermore, to evaluate the effectiveness of our algorithm on larger model and MoE architectures, we conduct experiments on Mistral-8$\times$7B. All experimental settings and datasets remain consistent with those in Section~\ref{subsec:setting}.

We compare CRVQ with AQLM. Experimental results demonstrate that, compared to the strong baseline AQLM, CRVQ consistently achieves superior perplexity and downstream task accuracy. This conclusion holds even for larger MoE models.

\begin{figure*}[t!]
    \centering
    \begin{subfigure}[b]{0.32\textwidth}
        \includegraphics[width=0.98\textwidth]{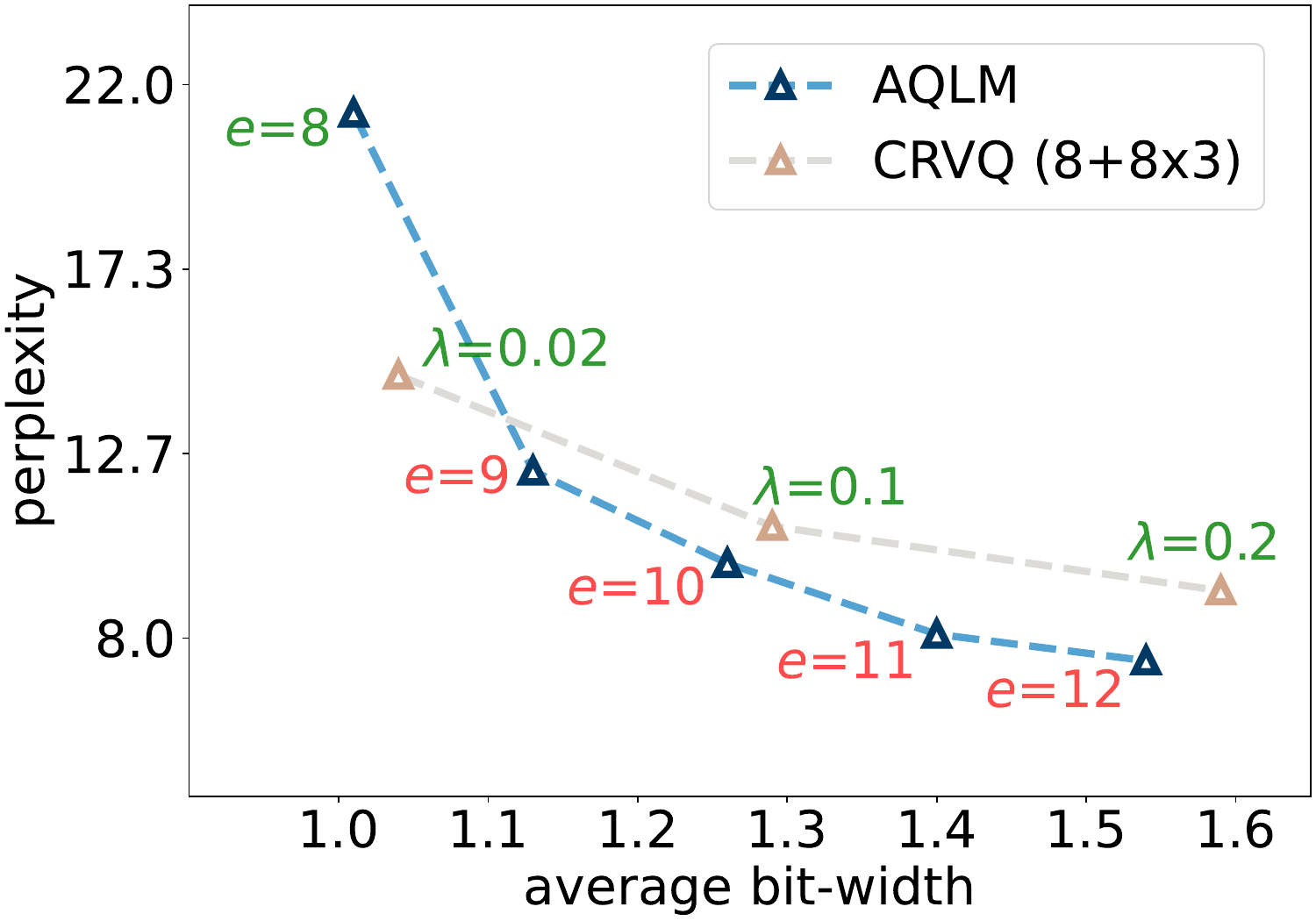}
        \caption{}
        \label{fig:scale_ppl}
    \end{subfigure}
    \begin{subfigure}[b]{0.32\textwidth}
        \includegraphics[width=0.98\textwidth]{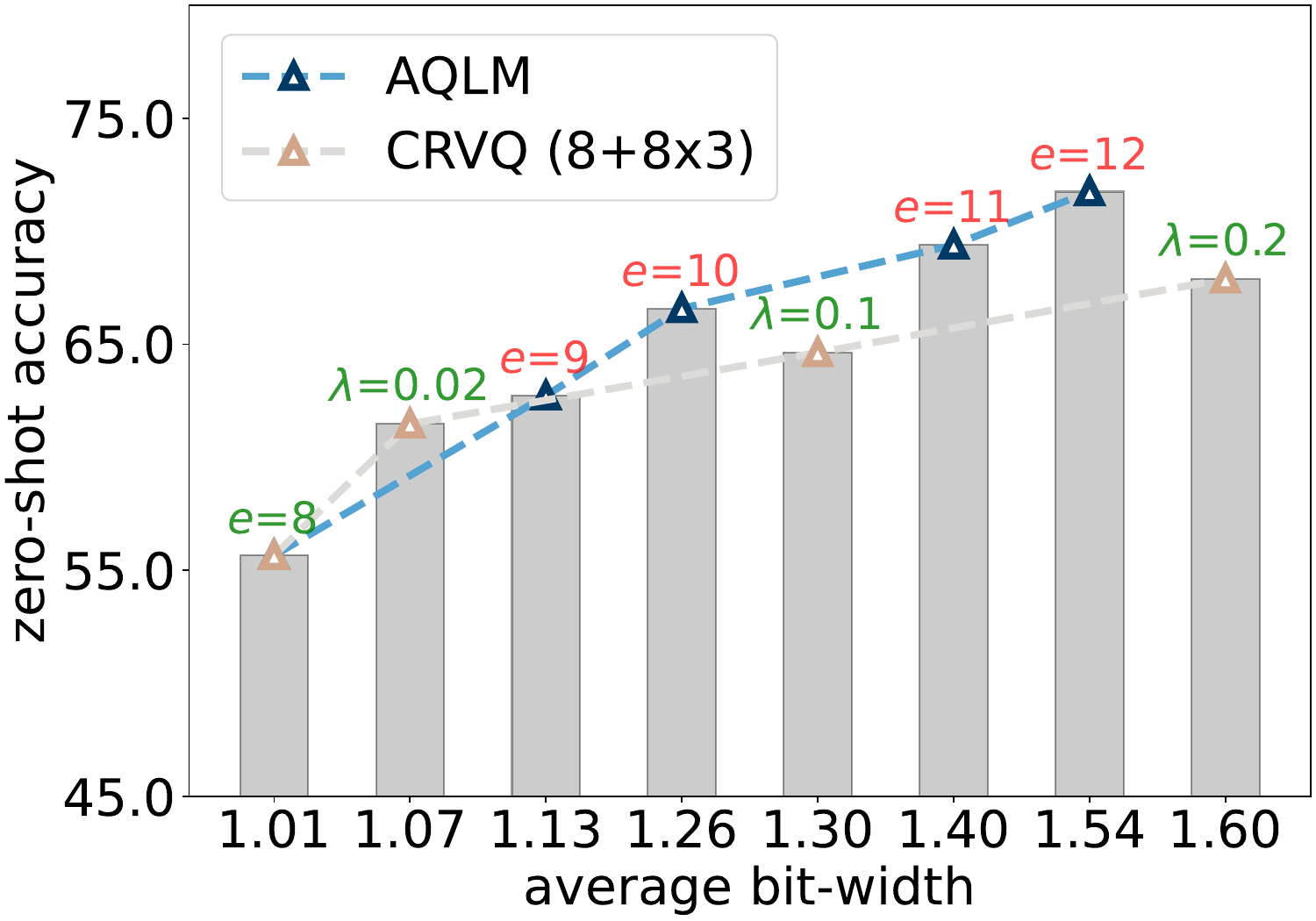}
        \caption{}
        \label{fig:scale_piqa}
    \end{subfigure}
    \begin{subfigure}[b]{0.32\textwidth}
        \includegraphics[width=0.98\textwidth]{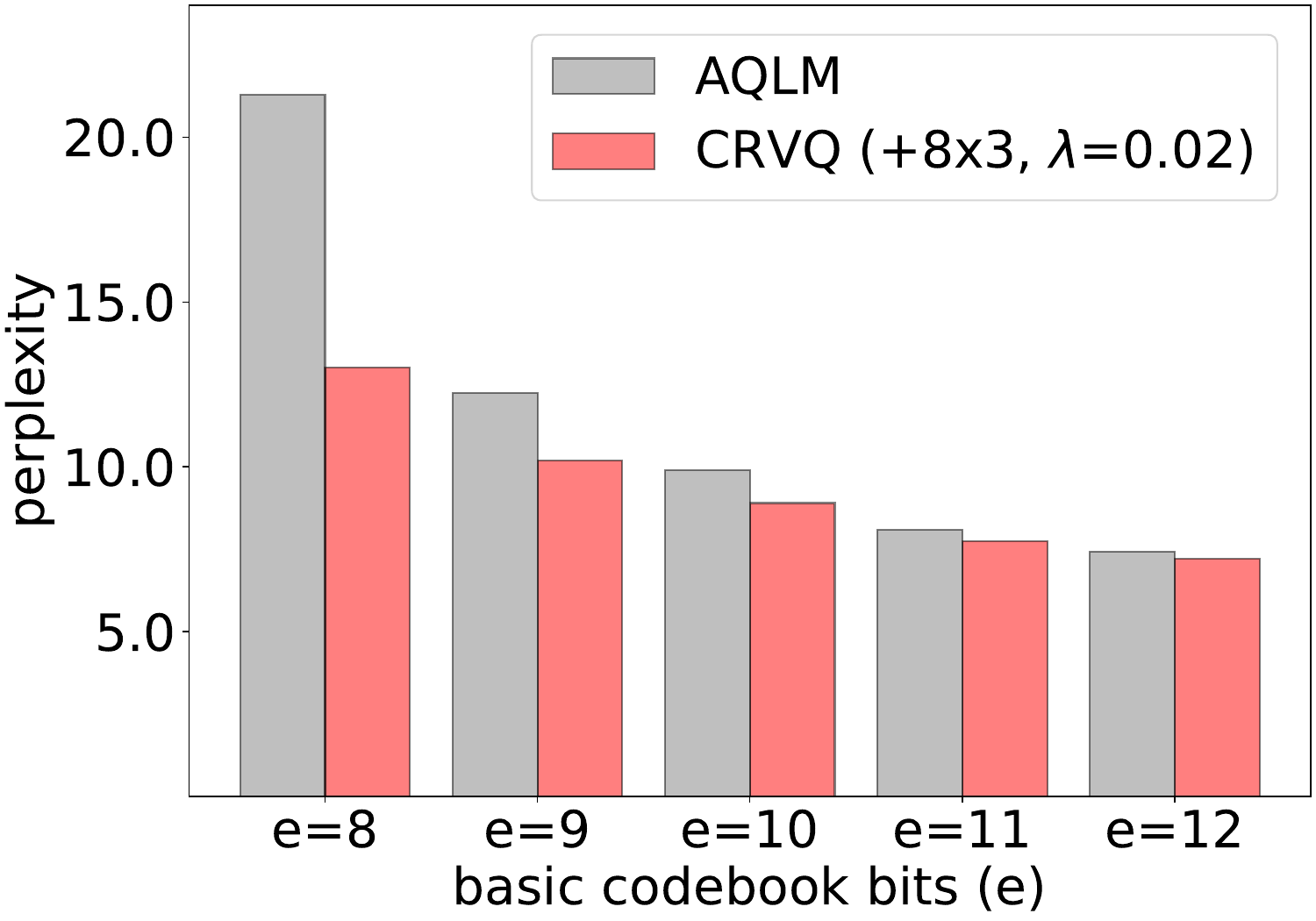}
        \caption{}
        \label{fig:scale_ppl2}
    \end{subfigure}
    \vspace{-0.5em}
    \caption{Bit scaling laws of AQLM and CRVQ on perplexity and zero-shot accuracy (PIQA). 
    In (a) and (b), we increase the AQLM codebook bit-width from 8 to 12 and compare it with CRVQ in different $\lambda$. Note that the basic codebook bit-width of CRVQ constantly equals 8 here. The hardware-unfriendly settings are noted in red. 
    In (c), the codebook bit-width of AQLM and basic codebook bit-width of CRVQ synchronously increase from 8 to 12.
    }
    \label{fig:scale}
    \vspace{-0.5em}
\end{figure*}

\begin{figure*}[t]
  \centering
  \begin{subfigure}[b]{\textwidth}
        \includegraphics[width=\textwidth]{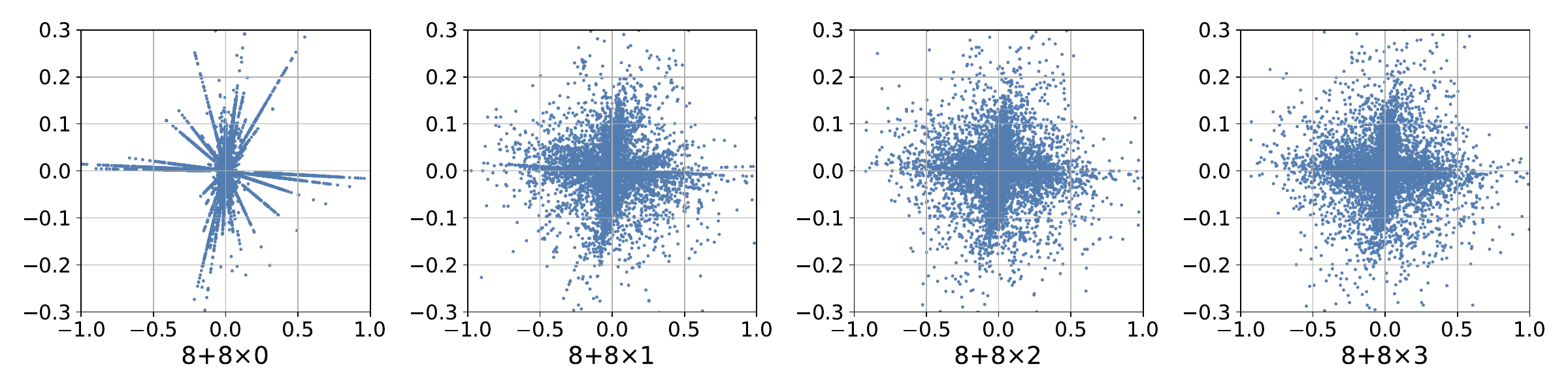}
        \vspace{-2.0em}
        \caption{vector samples from product of important weight channels and activation}
        \label{fig:vis_important}
  \end{subfigure}
  \begin{subfigure}[b]{\textwidth}
        \includegraphics[width=\textwidth]{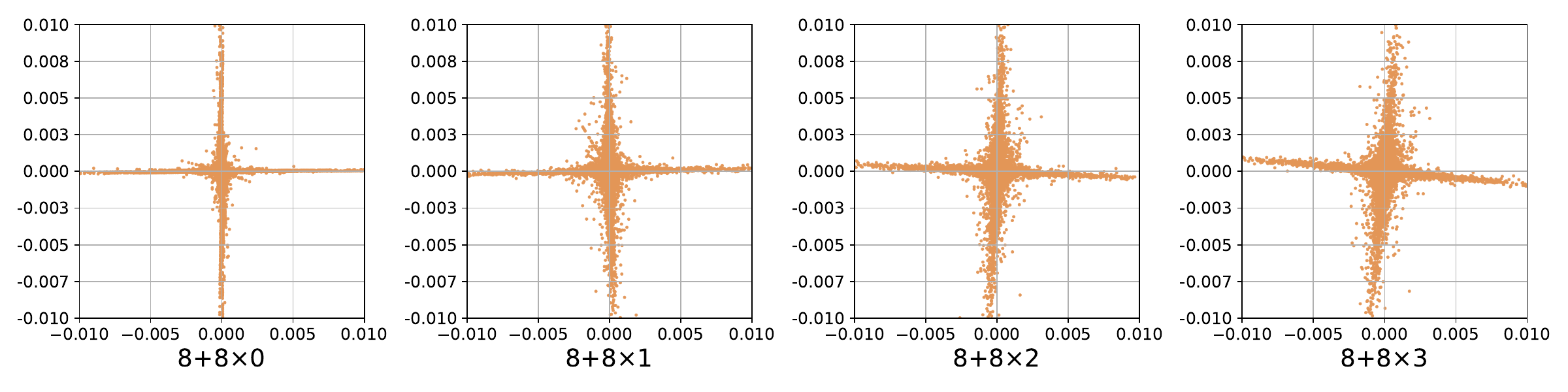}
        \vspace{-2.0em}
        \caption{vector samples from product of non-important weight channels and activation}
        \label{fig:vis_nonimportant}
  \end{subfigure}
  \vspace{-1.5em}
  \caption{2-D visualization of the element-wise product from quantized weight and activation. The weight matrix is \texttt{q\_proj} from layer 0 of LLaMA2-7B and we partition the product into vectors with a length of $d=8$. Here we use $\lambda=0.02$. (a) corresponds to important channels, while (b) corresponds to non-important channels. We compare the impact of different numbers of extended codebooks (ranging from 0 to 3) and channel importance on the distribution of product vectors.}
\label{fig:visual}
\vspace{-0.5em}
\end{figure*}

\subsection{Bit Scaling Laws}
\label{app:law}

In Section~\ref{subsec:deployment}, we discuss the device flexibility of CRVQ. To further illustrate its bit scaling property and deployment efficiency, we evaluate its perplexity and zero-shot accuracy under different bit-width configurations.

Compared to previous works, both AQLM and CRVQ allow flexible bit-width customization, enabling models to scale their capabilities according to hardware constraints. However, in extreme compression scenarios (sub-2-bit), AQLM can only adjust the codebook bit-width $e$ to control bit-width, as increasing the number of codebooks $m$ would double the total bit-width, exceeding 2. Yet, setting $e$ to 9, 10, 11, or 12 is impractical for deployment, as it misaligns memory boundary with hardware-friendly configurations, leading to inefficiencies and limited acceleration~\citep{aqlm2024}. This approach, even if effective, is not practical for real deployment (see Figure~\ref{fig:scale_ppl} and~\ref{fig:scale_piqa} for details).

In contrast, CRVQ offers a more deployment-friendly solution by keeping the basic codebook bit-width fixed at $e=8$ and adjusting the critical channel ratio $\lambda$ to achieve bit scaling, thereby enabling capability scaling. This method achieves similar effectiveness while maintaining a hardware-efficient computation process.

Additionally, if not considering hardware efficiency, we investigate how increasing the basic codebook bit-width in CRVQ affects its advantage over AQLM. In practice, as the basic codebook bit-width increases and model capacity improves, the advantage over AQLM gradually diminishes. This trend is illustrated in Figure~\ref{fig:scale_ppl2}. 
The reason for this lies in the application scenario of CRVQ, which focuses on exhuming bit-width efficiency in extreme compression. When the basic codebook provides sufficient ability, the contribution of critical channels becomes negligible.

\subsection{Codebook Fitting Visualization}
\label{app:visual}

We have demonstrated that a small subset of critical channels play a significant role in vector quantization. In our previous analysis, this is primarily reflected in overall LLM perplexity and downstream task accuracy. 
However, we remain curious about the differences between applying extended codebooks to important versus non-important channels. To figuratively understand this distinction, we partition the element-wise product of the quantized weight matrix and an activation vector into vectors of $d=8$, then apply PCA for 2-D visualization. The results from important and non-important channels are shown in Figure~\ref{fig:vis_important} and~\ref{fig:vis_nonimportant}, respectively.

From the Figure~\ref{fig:visual}, it is evident that the product vectors from important channels and non-important channels exhibit distinct distribution patterns after dimensionality reduction. The former resembles a scattered point cloud, while the latter forms a cross-like shape. 
As the number of extended codebooks increases, the product vectors of important channels gradually transition from a disordered spread to a cross-like structure, accompanied by a subtle rotational effect. In contrast, the product vectors of non-important channels primarily exhibit rotation without significant changes in shape as the number of extended codebooks increases.

These observations suggest that differences in channel importance manifest different quantization behaviors. Notably, the important weights change significantly with the increase in the number of extended codebooks.

\begin{table}[t]
\centering
\resizebox{0.99\columnwidth}{!}{
\renewcommand{\arraystretch}{1.0}
\begin{tabular}{l@{\hspace{0.40cm}}|c@{\hspace{0.30cm}}c@{\hspace{0.30cm}}|c@{\hspace{0.30cm}}c@{\hspace{0.30cm}}}
\toprule
\multirow{2}{*}[-0.8ex]{Model} & \multicolumn{2}{c|}{ARB-LLM} & \multicolumn{2}{c}{CRVQ} \\ 
\cmidrule(lr){2-3} \cmidrule(lr){4-5} 
& Wbits & Wiki2 & Wbits & Wiki2 \\ 
\midrule
\noalign{\vskip -1pt}
LLaMA-7B & 1.09 & 14.03 & 1.07 & \textbf{13.68} \\
LLaMA-13B & 1.09 & \textbf{10.18} & 1.06 & 10.73 \\
LLaMA2-7B & 1.08 & 16.44 & 1.07 & \textbf{13.01} \\
LLaMA2-13B & 1.08 & 11.85 & 1.06 & \textbf{9.81} \\
LLaMA3.1-8B & 1.06 & 27.42 & 1.07 & \textbf{27.36} \\
\noalign{\vskip -1pt}
\bottomrule
\end{tabular}
}
\caption{Comparison between ARB-LLM and CRVQ.}
\vspace{-1em}
\label{tab:arb}
\end{table}

\vspace{-1em}
\subsection{Comparison with ARB-LLM}
\label{app:arb}

A contemporaneous work, ARB-LLM~\citep{arb2024}, also explores 1-bit PTQ. It employs an alternating refined binarization approach to optimize the gap between binary and high-precision weights. While ARB-LLM also leverages the idea of certain weight sensitivity, it follows a different technical route from CRVQ, which is based on vector quantization. 
Table~\ref{tab:arb} compares the performance of both methods on LLaMA series models. Here CRVQ uses $8+8\times3$, $\lambda=0.02$, and $\textrm{ARB-LLM}_{\textrm{RC}}$ is reported. 
Although ARB-LLM has a clear advantage over CRVQ in terms of quantization runtime, CRVQ appears to achieve better performance.

\begin{table}[t]
\centering
\renewcommand{\arraystretch}{1.0}
\begin{tabular}{l@{\hspace{0.40cm}}|c@{\hspace{0.30cm}}c@{\hspace{0.30cm}}c@{\hspace{0.30cm}}}
\toprule
Method & 7B & 13B & 30B \\ 
\midrule
Float16 & 118$\mu s$ & 216$\mu s$ & 334$\mu s$ \\
AQLM $(8+8\times0)$ & $\times1.76$ & $\times1.97$ & $\times2.25$ \\
CRVQ $(8+8\times1)$ & $\times1.68$ & $\times1.96$ & $\times2.20$ \\
CRVQ $(8+8\times3)$ & $\times1.68$ & $\times1.93$ & $\times2.22$ \\
\bottomrule
\end{tabular}
\caption{Speed of the FP16 \texttt{gate\_proj} layer matrix-vector multiplication.}
\vspace{-0.5em}
\label{tab:speed}
\end{table}

\subsection{Inference Speed}
\label{app:speed}

We evaluate the computational performance of CRVQ on the same A100 GPU, with the results of LLaMA2 series presented in Table~\ref{tab:speed}. The experiments confirm that AQLM demonstrates the ability to accelerate matrix multiplication as they claimed. This is primarily because the benefits of I/O optimization outweigh the overhead introduced by dequantization. More importantly, since CRVQ focuses only on a small subset of critical weights, its additional computational cost is negligible. The limited use of extended codebooks does not lead to a distinct increase in time. These results highlight that CRVQ is highly friendly with deployment on resource-constrained devices.

\subsection{Details on Baselines and CRVQ}
\label{app:details}

In this subsection, we provide the essential supplemental‌ details of the baselines in this work and our proposed CRVQ:

\begin{itemize}
    \item PB-LLM~\citep{pbllm2024}: PB-LLM, which can be integrated with other classical PTQ algorithms, is an early approach to sub-2-bit PTQ, introduces mixed-precision quantization by applying higher bit-widths to a small subset of sensitive weights while binarizing the majority. We quantize the majority to 1-bit and maintain the sensitive weights to 8-bit, leading to a 1.7-bit algorithm. We use hessian metric to find the salient weights. The other parameters are based on the code released by the authors.
    \item BiLLM~\citep{billm2024}: BiLLM enhances quantization precision by partitioning the weight distribution and assigning different quantization parameters to each partition. It also employs residual-based multi-step quantization to improve accuracy. We use hessian metric to partition the weights and other settings are identical to the original paper.
    \item AQLM~\citep{aqlm2024}: AQLM is an additive vector quantization method that combines layer-level, block-level, and end-to-end fine-tuning. We limit the fine-tuning of intra-layer codebooks to a maximum of 100 epochs, typically not exceeding 20 epochs in most cases. For block-level and end-to-end fine-tuning, we perform a maximum of 25 epochs, using a batch size of 64. The fine-tuning process are optimized by Adam with $\beta_1=0.90, \beta_2=0.95$.
    \item CRVQ: The fine-tuning parameters in our method are the same as those used in AQLM. Given the use of multiple codebooks, we set the beam search width~\citep{aqlm2024} to 1, which is the same as the project AQLM~\citep{aqlm2024}.
\end{itemize}

\subsection{Supplemental Related Work}
\label{app:related}

Model quantization research includes several subfields. Based on the quantization object, it can be categorized into weight-only quantization~\citep{billm2024, quip2024, gptq2022} and weight-activation quantization~\citep{smooth2023, llmint2022}. Weight-only quantization (WxA16) focuses on low-bit weight representation. In contrast, weight-activation quantization (WxAy) quantizes both weights and intermediate activations. Model quantization can also be classified into quantization-aware training (QAT) and post-training quantization (PTQ) based on its integrating stage. 

As introduced in Section~\ref{sec:work}, this paper focuses on PTQ, and it also belongs to weight-only quantization. Due to space limitations, the survey and comparison of QAT are provided in the appendix.

QAT incorporates quantization into the training process, minimizing performance loss due to low precision and achieving superior results. LLM-QAT~\citep{llmqat2023} introduces a few learnable parameters and employs knowledge distillation to transfer abilities from the original model. QLoRA~\citep{qlora2023} mitigates the adverse effects of quantization by fine-tuning high-precision low-rank adapters. Recently, researchers explore QAT for more effective extreme compression. EfficientQAT~\citep{efqat2024} adjusts all model through a novel block-wise fine-tuning, achieving outstanding results at the 2-bit quantization level while reducing computational overhead. OneBit~\citep{onebit2024} designs a new architecture for 1-bit linear layers, incorporating two high-precision adapter vectors and leveraging knowledge distillation to enable effective 1-bit quantization. BinaryMoS~\citep{bmos2024} further improves on OneBit by replacing the vectors with a mixture-of-experts paradigm. Although QAT is effective for extreme quantization, it still faces challenges such as performance constraints and substantial computational costs. We provide comparison between our CRVQ (PTQ) and OneBit (QAT) in Section~\ref{subsec:qat}.

\subsection{Generation Cases}
\label{app:cases}

We quantize the LLaMA2-13B model using different methods and compare their generated outputs for given prompts, as shown in Table~\ref{tab:cases}. The average bit-width of the models is 1.08, 1.01, and 1.06 bits respectively. The results indicate that CRVQ produces longer and more fluent text. However, despite its fluency, the generated content still contains many hallucinations. For example, in case 3, the capital of New York State is 150 miles from New York City, not 1,100 miles, and there are 62 counties, not 10.

\begin{table*}[t!]
  \centering
  \scalebox{0.98}{
  \begin{tabular}{c m{14cm}}
  \toprule
  \textcolor{blue}{\bf Prompt} & \textcolor{red}{A cowboy rides a\_} \\
  \cmidrule(lr){1-1} \cmidrule(lr){2-2}
  \bf BiLLM & cowboy rides a cowboy \\
  \cmidrule(lr){1-1} \cmidrule(lr){2-2}
  \bf AQLM & troller. in the troller. The troller is a troller. \\
  \cmidrule(lr){1-1} \cmidrule(lr){2-2}
  \bf CRVQ & 1950s Ford Mustang in the desert. The cowboy rides a 1950s Ford Mustang in the desert. \\
  \midrule
  \textcolor{blue}{\bf Prompt} & \textcolor{red}{There are a number of ways to reduce air pollution, such as\_} \\
  \cmidrule(lr){1-1} \cmidrule(lr){2-2}
  \bf BiLLM & 100\% of the air pollution. The air pollution is a number of the air pollution. \\
  \cmidrule(lr){1-1} \cmidrule(lr){2-2}
  \bf AQLM & 10000000000000000000000 \\
  \cmidrule(lr){1-1} \cmidrule(lr){2-2}
  \bf CRVQ & 1) reducing the amount of pollution from the burning of fossil fuels, 2) reducing the amount of pollution from the burning of coal, 3) reducing the amount of pollution from the burning of natural gas, 4) reducing the amount of pollution from the burning of wood, 5) reducing the amount of pollution from the burning of natural gas \\
  \midrule
  \textcolor{blue}{\bf Prompt} & \textcolor{red}{The capital of the state of New York is\_} \\
  \cmidrule(lr){1-1} \cmidrule(lr){2-2}
  \bf BiLLM & 100\% of the state of New York is 100\% of the state of New York \\
  \cmidrule(lr){1-1} \cmidrule(lr){2-2}
  \bf AQLM & 100\% of the state of New York. of the state of New York \\
  \cmidrule(lr){1-1} \cmidrule(lr){2-2}
  \bf CRVQ & 1100 miles away from the city of New York. The state of New York is divided into 10 counties. \\
  \bottomrule
  \end{tabular}}
  \caption{Generation cases of CRVQ and baselines. The three prompt we use is from~\citep{bmos2024}.}
  \label{tab:cases}
\end{table*}

\end{document}